\documentclass[format=acmsmall,screen=true,noacm]{acmart}

\AtBeginDocument{%
  }

\acmJournal{THRI}
\acmVolume{37} % to be updated
\acmNumber{4} % to be updated
\acmArticle{111} % to be updated
\acmMonth{8} % to be updated

\usepackage{booktabs}
\graphicspath{{Figures}}
\DeclareGraphicsExtensions{.png}
\usepackage{bigstrut}
\usepackage{multirow}
\usepackage{multicol}

\newcommand{\p}[1]{\smallskip \noindent \textbf{{#1}.}}
\newcommand{\eq}[1]{Equation~(\ref{eq:#1})}
\newcommand{\fig}[1]{Figure~\ref{fig:#1}}

\usepackage{graphicx}
\graphicspath{{Figures}}
\DeclareGraphicsExtensions{.png}

\usepackage{enumitem}
\usepackage{algorithm, setspace}
\usepackage{algpseudocode}
\usepackage{csquotes}
\usepackage{balance}
\usepackage{amsmath}
\usepackage{bbm}

\setcopyright{none}

\begin{document}

%%%%%%%%%%%%%%%%%%%%%%%%%%%%%%%%%%%%%%%%%%%%%%%%%%%%%%%%%%%%%%%%%%%%%%%%%%%%%%%%
\title{LIMIT: Learning Interfaces to Maximize Information Transfer}

%%%%%%%%%%%%%%%%%%%%%%%%%%%%%%%%%%%%%%%%%%%%%%%%%%%%%%%%%%%%%%%%%%%%%%%%%%%%%%%%

\author{Benjamin A. Christie}
\orcid{0000-0003-1918-0892}
\affiliation{
  \institution{Virginia Tech}
  \department{Department of Mechanical Engineering}
  \streetaddress{635 Prices Fork Rd}
  \city{Blacksburg}
  \state{VA}
  \postcode{24060}
  \country{USA}
  }
\email{benc00@vt.edu}

\author{Dylan P. Losey}
\orcid{0000-0002-8787-5293}
\affiliation{
  \institution{Virginia Tech}
  \department{Department of Mechanical Engineering}
  \streetaddress{635 Prices Fork Rd}
  \city{Blacksburg}
  \state{VA}
  \postcode{24060}
  \country{USA}
  }
\email{losey@vt.edu}

\thanks{This work is supported in part by NSF Grant \#2129201.}

%%%%%%%%%%%%%%%%%%%%%%%%%%%%%%%%%%%%%%%%%%%%%%%%%%%%%%%%%%%%%%%%%%%%%%%%%%%%%%%%

\begin{abstract}

Robots can use auditory, visual, or haptic interfaces to convey information to human users. The way these interfaces select signals is typically pre-defined by the designer: for instance, a haptic wristband might vibrate when the robot is moving and squeeze when the robot stops. But different people interpret the same signals in different ways, so that what makes sense to one person might be confusing or unintuitive to another. In this paper we introduce a unified algorithmic formalism for learning \textit{co-adaptive} interfaces from \textit{scratch}. Our method does not need to know the human's task (i.e., what the human is using these signals for). Instead, our insight is that interpretable interfaces should select signals that maximize \textit{correlation} between the human's actions and the information the interface is trying to convey. Applying this insight we develop LIMIT: Learning Interfaces to Maximize Information Transfer. LIMIT optimizes a tractable, real-time proxy of information gain in continuous spaces. The first time a person works with our system the signals may appear random; but over repeated interactions the interface learns a one-to-one mapping between displayed signals and human responses. Our resulting approach is both personalized to the current user and not tied to any specific interface modality. We compare LIMIT to state-of-the-art baselines across controlled simulations, an online survey, and an in-person user study with auditory, visual, and haptic interfaces. Overall, our results suggest that LIMIT learns interfaces that enable users to complete the task more quickly and efficiently, and users subjectively prefer LIMIT to the alternatives. See videos here: \url{https://youtu.be/IvQ3TM1_2fA}.

% Robots can use auditory, visual, or haptic interfaces to convey information to human users. The way these interfaces select signals is typically pre-defined by the designer: for instance, a haptic wristband might vibrate when the robot is moving and squeeze when the robot stops. But different people interpret the same signals in different ways, so that what makes sense to one person is confusing to another. In this paper we introduce a unified algorithmic formalism for learning co-adaptive interfaces from scratch. Our insight is that interpretable interfaces should select signals that maximize correlation between the human's actions and the information the interface is trying to convey. Applying this insight we develop LIMIT: Learning Interfaces to Maximize Information Transfer. LIMIT optimizes a tractable, real-time proxy of information gain in continuous spaces. The first time a person works with our system the signals may appear random; but over repeated interactions the interface learns a one-to-one mapping between signals and human responses. Our resulting approach is both personalized to the current user and not tied to any specific interface modality. We compare LIMIT to state-of-the-art baselines across controlled simulations, an online survey, and an in-person user study with auditory, visual, and haptic interfaces.

\end{abstract}






%%%%%%%%%%%%%%%%%%%%%%%%%%%%%%%%%%%%%%%%%%%%%%%%%%%%%%%%%%%%%%%%%%%%%%%%%%%%%%%%

\setcopyright{rightsretained}
\acmJournal{THRI}
\acmYear{2024} \acmVolume{1} \acmNumber{1} \acmArticle{1} \acmMonth{1}\acmDOI{10.1145/3675758}

\maketitle

\section{Introduction}

\begin{figure}[t]
	\begin{center}
		\includegraphics[width=0.7\columnwidth]{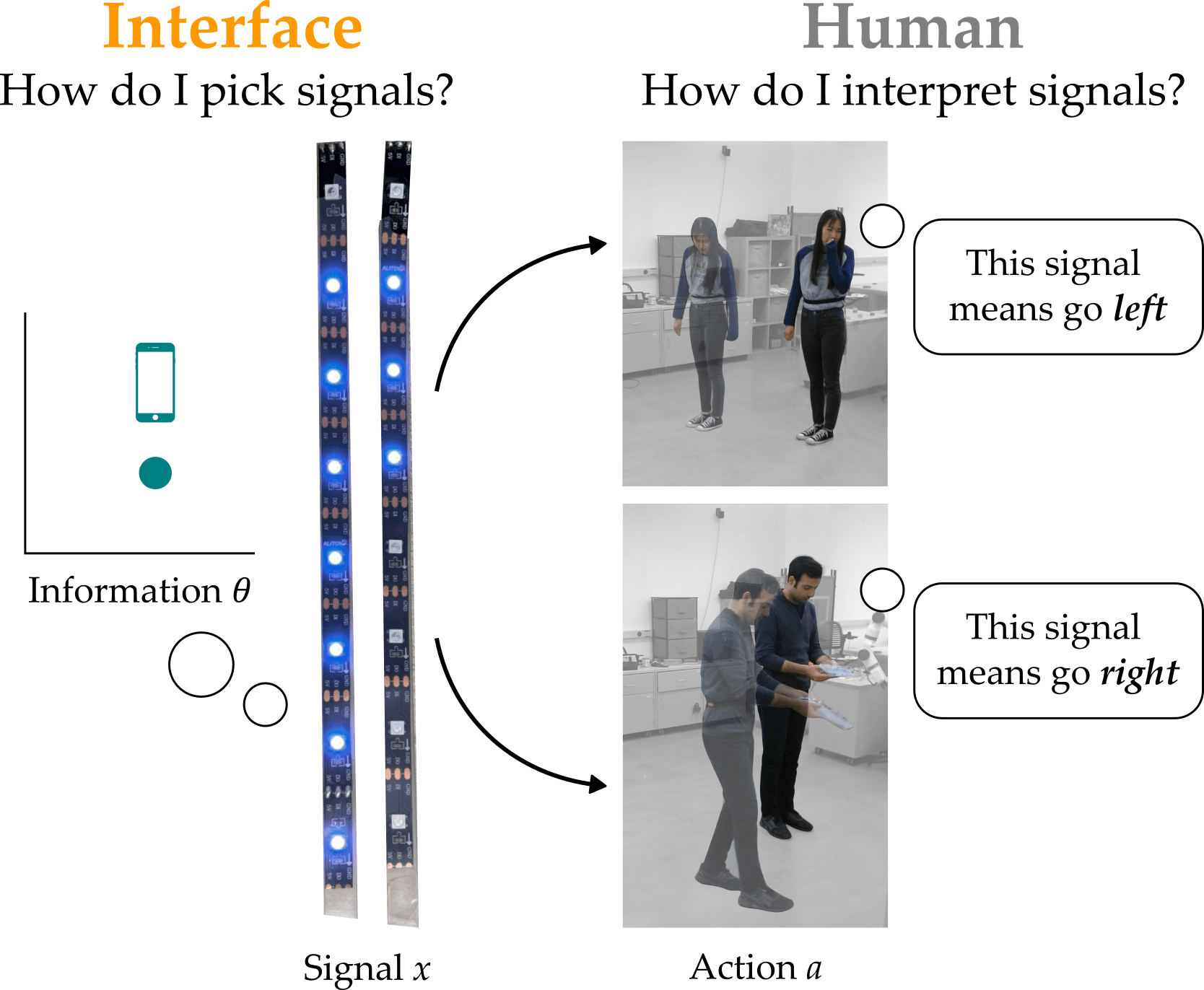}
		\caption{Interface selecting signals to convey information to the human operator. Choosing the right feedback is challenging because the way people respond to signals varies across tasks, users, and interface types; e.g., when a person sees this LED pattern should they go left or right? We introduce a unified algorithmic framework that co-adapts to the current user by learning to pick signals that maximize information transfer.}
		\label{fig:front}
	\end{center}
\end{figure}

Imagine a person collaborating with a robotic interface to complete some task. The interface displays signals to convey information to the person, and the person interprets those signals to determine what actions to take. For instance, in \fig{front} the human is searching for their missing phone. The interface knows the phone's location and can signal the human with an array of LEDs. But how does the interface determine which signals to use? One person might think that the left LED strip corresponds to the phone's position in the $x$-axis and the right strip indicates position in the $y$-axis. But another user might have the opposite mapping --- or interpret the interface's feedback in an entirely different way. For each user, the interface must identify a method for selecting signals (e.g., turning on LEDs) that clearly conveys the desired information (e.g., the phone's location).

In this paper we explore settings where a robotic interface is communicating information to a human operator. Here \textit{interfaces} refer to autonomous systems that provide nonverbal feedback in the form of lights, sounds, augmented reality displays, haptic signals, or robot motion. We assume that the interface has access to some task-related, \textit{hidden} information that the human cannot directly observe. Existing research pre-programs these interfaces with a human-engineered and fixed mapping from information to signals \cite{hellstrom2018understandable, cha2018survey}. Returning to our motivating example, state-of-the-art methods might tell the robot to always use the left LEDs for $x$-position and the right LEDs for $y$-position. But there are two fundamental limitations of this approach. First, using a fixed convention for choosing signals \textit{forces} all humans to learn and follow this specific convention; by contrast, we know that humans have personalized signal preferences and interpretations \cite{belpaeme2018social, tucker2020preference, gasteiger2021factors}. Second, these human-engineered mappings must be designed on a \textit{case-by-case} basis, where the designers rely on their intuition and experimental data to decide how the interface will provide feedback for the current task \cite{seifi2020novice, szafir2021connecting, suzuki2022augmented}.

To overcome these limitations we here introduce a \textit{unified} algorithmic framework for \textit{learning} interfaces from scratch. We do not assume that the interface (a) knows the task the human wants to complete or (b) has a model of how the human will interpret its signals. Instead, our insight is that ---  in tasks where a robotic interface is sharing \textit{hidden} information with a user:
\begin{center}
\textit{When interfaces are \emph{interpretable} the human's actions are \emph{correlated} \\ with the hidden information that the interface is trying to convey}.
\end{center}

Effective feedback signals should guide the user’s decisions and inform the human’s behaviors. Return to our motivating example where the robotic interface is trying to communicate the position of the human's missing phone. Applied to this example setting, our insight asserts that --- if the phone's location changes --- the interface should display LED signals that cause the human's actions to also change. For instance, if the phone is on the left side of the room the interface should display different signals (and cause the human to take different actions) as compared to when the phone is on the right side of the room.

We use this insight to develop LIMIT: \textbf{L}earning \textbf{I}nterfaces to \textbf{M}aximize \textbf{I}nformation \textbf{T}ransfer. LIMIT is an information-theoretic algorithm that learns a real-time interface policy (i.e., a mapping from information to signals) to maximize the mutual dependence between the hidden information and the human actions. LIMIT is not tied to any specific type of interface; as we will show, our unified approach can be applied to visual displays, auditory cues, and haptic arrays. The first time a new user interacts with LIMIT the interface signals may appear random or irregular. But over repeated interactions LIMIT gathers data from the current user, learns online, and personalizes the signals so that humans take different actions for different values of hidden information. Co-adaptive humans exploit these interpretable signals to complete the task and maximize their reward over repeated interactions. Overall, LIMIT is a step towards robots that learn how to convey their own latent, internal information to nearby humans.

In this paper we make the following contributions:

\p{Formalizing Interfaces with Information Transfer} We formulate robotic interfaces as the intersection of human and interface policies. In settings where an interface has access to hidden information that a human does not, we hypothesize that interpretable, task-agnostic interfaces should maximize conditional information gain between hidden information and human actions. We then derive information gain in terms of the agent policies.

\p{Learning Interfaces to Maximize Information Gain} Directly optimizing for information gain is intractable in continuous spaces. We accordingly introduce LIMIT, an online learning approach that closely mirrors the structure of our derived formulation\footnote{Our code for implementing LIMIT is available here: \url{https://github.com/VT-Collab/LIMIT-learning-interfaces}}. LIMIT learns an interface policy to correlate the human's actions and hidden information.

\p{Comparing Interfaces in Controlled Simulations} We compare LIMIT to ablations and a state-of-the-art baseline across simulated environments. This includes settings where the interface signal is over-actuated (i.e., the signal has more dimensions than the information) and under-actuated (i.e., the information has more dimensions than the signal).

\p{Testing Interface Interpretability with Online Users} We perform an online user study where $37$ participants attempt to find their missing phone using learned interface feedback. Participants more accurately completed the task with LIMIT feedback, and also perceived the LIMIT interface as more helpful, understandable, and intuitive.

\p{Conducting User Studies on Visual, Auditory, and Haptic Interfaces} We put LIMIT to the test with $11$ in-person users across three different types of interfaces. Participants guide a robot arm or walk around a room while getting sound, light, and haptic feedback. In each task the interface must co-adapt alongside the human and learn to select meaningful signals. As compared to a state-of-the-art baseline, LIMIT results in better objective performance and subjective ratings.
\section{Related Work} \label{sec:related}

We focus on communicating information from robotic interfaces to human operators. The interface knows some information and needs to determine what signals it should display to convey that information to the human. Our goal is to develop a \textit{unified} algorithmic framework that can be applied to different tasks and types of interfaces, and learns to output interpretable signals that are personalized to the current user. Here we discuss related research that leverages fixed, pre-defined interface mappings, as well as interfaces that learn to interpret the human's inputs.

\p{Pre-Defined Interfaces} Prior works explore how interfaces can convey information to humans through nonverbal cues such as lights, projections, augmented reality, haptic signals, and robot motion \cite{cha2018survey}. Often the interfaces are designed with a specific task in mind, and programmed with a \textit{pre-defined} mapping from information to signals that is held \textit{constant} throughout human-interface interaction \cite{chadalavada2015s, weng2019robot, andersen2016projecting, walker2018communicating, rosen2019communicating, mullen2021communicating, tan2020acquisition, dunkelberger2018conveying, che2020efficient}. The resulting signals can be intuitive for users to interpret without much experience or explanation \cite{chadalavada2015s, weng2019robot, andersen2016projecting, walker2018communicating, rosen2019communicating, mullen2021communicating}. For example, in \cite{chadalavada2015s} a mobile robot drove around a crowded room; when the robot projected a line onto the floor indicating its planned trajectory, humans quickly recognized the robot's intent. In other settings the fixed, hand-designed mappings are high-dimensional or intricate, and users may need practice to correctly interpret the robot's meaning \cite{tan2020acquisition, dunkelberger2018conveying, che2020efficient}. For instance, in \cite{tan2020acquisition} humans were trained to convert tactile signals into $500$ different words (similar to Morse code). Rather than pre-defining the mapping from information to signals, alternate research assumes the interface has an accurate \textit{model of the human} \cite{dragan2013legibility, kwon2018expressing, huang2019enabling}. More specifically, these works assume that the robot knows how the human will interpret its motions; the robot then inverts this model to select legible behaviors and convey the desired information. This approach works well when the actual user follows the robot's convention --- but we know that different humans will interpret and respond to the same feedback in different ways \cite{belpaeme2018social, tucker2020preference, gasteiger2021factors}. Unlike these prior works we do not assume that the interface is given either a pre-defined mapping from information to signals or a human model. Instead, we seek to \textit{learn} an interpretable and personalized mapping from scratch.

\p{Learned Interfaces} Most relevant here is recent research that learns mappings from the \textit{human inputs} (i.e., signals) to the \textit{human's intent} (i.e., information) \cite{reddy2022first, li2020learning, de2021framework, rizzoglio2021building}. For instance, in \cite{reddy2022first} the human is controlling a drone with a keyboard, and the robot learns how to map the human's key presses to drone motions. Similarly, in \cite{li2020learning} a human is controlling an assistive robot arm with a joystick, and the robot learns which joystick directions should be associated with each robot motion. These learned mappings go in the \textit{opposite} of what we are interested in: instead of learning how to extract information from human commands, we want to learn how to convey the interface's information to human operators. Put another way, in our work the interface is sending signals to the human. As the interface learns from and adapts to the human operator, the human will inevitably co-adapt to the interface \cite{nikolaidis2017human, parekh2022learning, ikemoto2012physical, van2021becoming}. Building on this prior work, we recognize that humans are not static operators: our approach must be able update and refine signals as the human learns how to interpret the interface.

\p{Maximizing Information Gain} Under our proposed approach the interface learns to display signals that maximize the correlation between the human's actions and the interface's information. More specifically, we will develop an algorithm where the interface learns to maximize a proxy of \textit{information gain}. Recent works have similarly leveraged information gain (i.e., mutual information) to select robot behaviors during human-robot interaction \cite{reddy2022first, jaques2019social, sadigh2018planning, habibian2022here, lee2021pebble, kaupp2010human}. For example, in \cite{sadigh2018planning} an autonomous car nudges closer to the human's car to see how the human will respond (and actively gather information about the human driver). Likewise, in \cite{kaupp2010human} a social robot communicates with the human when the expected benefits of the human's feedback outweigh the cost of probing the human. Although our proposed approach similarly optimizes for information gain, we do so in the opposite direction: the related works \cite{jaques2019social, sadigh2018planning, habibian2022here, lee2021pebble, kaupp2010human} select robot actions to \textit{gain} information from the human, while we will choose interface signals to \textit{convey} information to the human.
\section{Problem Formulation} \label{sec:problem}

We consider settings where a feedback interface is sending signals to a human. Our approach is not tied to any specific type of interface: e.g., the interface could be a haptic wristband, a light projection, or an augmented reality display. The human is attempting to perform some task. We assume that the interface knows \textit{hidden information} $\theta$ that the human cannot directly observe, {and the human's task depends on this hidden information. More specifically, we assume that the human should take different actions if the hidden information changes.} In our running example (see \fig{front}) the interface is a wearable array of LED lights. Here the human's task is to find their phone: only the interface knows the phone's location $\theta$, and the human must interpret the feedback signals to reach $\theta$. Our fundamental challenge is finding a signal mapping that is \textit{interpretable} for the current user. We do not assume that the human and interface have a pre-defined convention for signals (e.g., the human and robot do not assume that the left lights indicate horizontal motion and the right lights indicate vertical motion). Instead, we want to enable interfaces to learn to communicate with the current user from scratch.

\p{Human} Let $s \in \mathcal{S}$ be the system state and let $a \in \mathcal{A}$ be the human's action. In our running example the state is the position of the human and the human's action is their change in position. The state transitions based on the human's action:
\begin{equation} \label{eq:P1}
    s^{t+1} = f(s^t, a^t)
\end{equation}
where $f$ is the deterministic dynamics and $t$ is the current timestep. An interaction lasts a total of $T$ timesteps. We use trajectory $\xi = (s^1, s^2, \ldots, s^T)$ to capture the sequence of states the human visits during the current interaction.

\p{Interface} Let $\theta \in \Theta$ be the hidden information and let $P(\theta)$ be a prior over this information. In our running example $\theta$ is the phone's location and $P(\theta)$ is a uniform distribution across the room. At the start of the interaction the interface observes $\theta \sim P( \cdot )$, and this parameter remains constant throughout the rest of the interaction. At each timestep $t$ the interface sends signal $x \in \mathcal{X}$ to the human, where $\mathcal{X}$ is the set of all possible signals the interface can output. For our running example $x$ is the intensity of the LED light array.

\p{Policies} The interface observes the system state $s$ and hidden information $\theta$ and then outputs signal $x$. Accordingly, the interface's policy maps $(s, \theta)$ to $x$:
\begin{equation} \label{eq:P2}
    \pi_\mathcal{R}(x \mid s, \theta)
\end{equation}
The human sees states and signals; importantly, the human \textit{cannot} directly observe the hidden information $\theta$. We assume the human gets the current signal $x^t$ before taking action $a^t$, so that the human's policy is a mapping from states and signals to actions:
\begin{equation} \label{eq:P3}
    \pi_\mathcal{H}(a \mid s, x)
\end{equation}

\p{Objective} During each interaction the human has in mind some task that they want to complete. This task depends on the hidden information $\theta$; for instance, perhaps the human wants to locate their phone and $\theta$ is the phone's position. Let the human have reward function $R(\xi, \theta) \rightarrow \mathbb{R}$, where higher rewards indicate that the human has better accomplished their task. We \textit{do not assume} that the interface has any knowledge of this task or reward function. Instead, the interface only has access to the data it has directly observed: the states, actions, signals, and hidden information from previous interactions. Based only on this data, we seek to learn an interface policy $\pi_\mathcal{R}$ that --- when paired with the human policy $\pi_\mathcal{H}$ --- will maximize the human's reward $R(\xi, \theta)$. 
\section{Learning Interfaces to Maximize Information Transfer (LIMIT)} \label{sec:method}

Given an interface and hidden information $\theta$, our goal is to find an interface policy $\pi_\mathcal{R}$ that helps the human complete their task and maximize their reward. This is challenging because (a) the interface does not know the human's task and (b) different humans respond to the same signals in different ways. Within this section we accordingly develop a \textit{task-agnostic}, \textit{personalized} approach for learning interface mappings. This approach is based on our fundamental insight that the human's actions should be correlated with the information that the interface is trying to communicate. In Section~\ref{sec:M1} we capture the mutual dependence between human actions and hidden information using conditional \textit{information gain}. We then rewrite this information gain in terms of the human and interface policies (Section~\ref{sec:M2}). Using these equations we introduce a real-time learning approach that trains the interface policy to personalize to the human's current behavior (Section~\ref{sec:M3}). Finally, in Section~\ref{sec:M4} we account for the human's co-adaptation to the changing interface.

\subsection{Optimizing for Information Gain} \label{sec:M1}

Information gain (i.e., mutual information) is a general metric for correlation: it quantifies the amount of information obtained about one variable by observing another variable \cite{cover2012elements}. Our central hypothesis is that an effective interface should maximize the correlation between human actions and hidden information. Accordingly, we assert that the interface should maximize the \textit{conditional information gain} between action $a$ and hidden information $\theta$ given state $s$:
\begin{equation} \label{eq:M1}
   I(a ~;~ \theta \mid s) = H(a \mid s) - H(a \mid s, \theta)
\end{equation}
Here $H(a \mid s)$ is the conditional Shannon entropy of $a$ given $s$, and $H(a \mid s, \theta)$ is the conditional Shannon entropy of $a$ given $s$ and $\theta$. Intuitively, $H(a \mid s)$ captures how uncertain we are about the human's action at state $s$, while $H(a \mid s, \theta)$ captures our uncertainty given both $s$ and $\theta$. 

\eq{M1} is maximized when (a) each action is equally likely at state $s$ but (b) we know exactly which action the human will take once hidden information $\theta$ is observed. Consider our running example where a human is standing in the middle of the room. The human could walk in any direction; but once the human knows their phone's location $\theta$, the human goes directly towards that goal. More generally, an interface that maximizes \eq{M1} will cause the human operator to take actions $a$ that are correlated with the hidden information $\theta$ the interface is trying to convey.

\subsection{Writing Information Gain in Terms of Policies} \label{sec:M2}

We want to learn an interface policy $\pi_\mathcal{R}$ that optimizes the conditional information gain $I(a; \theta \mid s)$. Towards this end, we here rewrite \eq{M1} in terms of the human policy $\pi_\mathcal{H}$ and the interface policy $\pi_\mathcal{R}$. For ease of explanation we treat $\mathcal{S}$, $\mathcal{A}$, $\mathcal{X}$, and $\Theta$ as discrete sets: the same result extends to continuous spaces by replacing the following summations with integrals over continuous distributions. We will work in continuous spaces during our simulations and user study.

By definition \eq{M1} is equal to \cite{cover2012elements}:
\begin{equation} \label{eq:M2}
    I(a ~;~ \theta \mid s) = \sum\limits_{\mathcal{S}, \mathcal{A}, \Theta} P(s, a, \theta) \log \frac{P(a \mid \theta, s)}{P(a \mid s)}
\end{equation}
Marginalizing over the interface signal $x$ at each term we find that:
\begin{equation} \label{eq:M3}
    I(a ~;~ \theta \mid s) = \sum\limits_{\mathcal{S}, \mathcal{A}, \Theta} \bigg(\sum_\mathcal{X}P(s, a, x, \theta)\bigg) \log \frac{\sum_\mathcal{X} P(a, x \mid \theta, s)}{\sum_\mathcal{X} P(a, x \mid s)}
\end{equation}
Remember that the human and interface policies are probability distributions: $\pi_\mathcal{R}$ is the probability of signal $x$ given $s$ and $\theta$, and $\pi_\mathcal{H}$ is the probability of action $a$ given $s$ and $x$. Because \eq{P3} only depends on $s$ and $x$, we further have that $P(a \mid s, x, \theta) = \pi_\mathcal{H}(a \mid s, x)$. Using the chain rule and plugging in \eq{P2} and \eq{P3}, we get that $I(a ~;~ \theta \mid s)$ from \eq{M1} is equal to:
\begin{equation} \label{eq:M4}
    \sum_{\mathcal{S}, \mathcal{A}, \Theta} P(s, \theta) \bigg(\sum_\mathcal{X}\pi_\mathcal{H}(a \mid s, x) \cdot \pi_\mathcal{R}(x \mid s, \theta)\bigg) \cdot \log \frac{\sum_\mathcal{X} \pi_\mathcal{H}(a \mid s, x) \cdot \pi_\mathcal{R}(x \mid s, \theta)}{\sum_\mathcal{X} \pi_\mathcal{H}(a \mid s, x) \sum_{\Theta} \pi_\mathcal{R}(x \mid s, \theta') P(\theta')}
\end{equation}
\eq{M4} re-expresses conditional mutual information in terms of the human policy $\pi_\mathcal{H}$ and interface policy $\pi_\mathcal{R}$.

We will gain additional insight by separating \eq{M4} into two terms: \textit{convey} and \textit{distinguish}. We refer to the first term as \textit{convey}:
\begin{equation} \label{eq:M5}
    \mathcal{T}_{conv} = \sum_\mathcal{X}\pi_\mathcal{H}(a \mid s, x) \cdot \pi_\mathcal{R}(x \mid s, \theta)
\end{equation}
Note that $\mathcal{T}_{conv}$ appears twice in \eq{M4}: once outside of the $\log$ and again in the numerator of the $\log$. Next, we refer to our second term as \textit{distinguish}:
\begin{equation} \label{eq:M6}
    \mathcal{T}_{dist} = \sum_\mathcal{X} \pi_\mathcal{H}(a \mid s, x) \sum_{\Theta} \pi_\mathcal{R}(x \mid s, \theta') P(\theta')
\end{equation}
For clarity, we show how these \textit{convey} and \textit{distinguish} terms are derived from \eq{M4} below:
\begin{equation} \label{eq:M7}
    I(a ~;~ \theta \mid s) = \sum_{\mathcal{S}, \mathcal{A}, \Theta} P(s, \theta) \cdot \mathcal{T}_{conv} \cdot \log \frac{\mathcal{T}_{conv}}{\mathcal{T}_{dist}}
\end{equation}
$\mathcal{T}_{conv}$ captures how likely it is that the human takes action $a$ given state $s$ and hidden information $\theta$. By contrast, $\mathcal{T}_{dist}$ expresses the likelihood of action $a$ at the current state across any choice of $\theta$. From \eq{M7}, we see that an interface $\pi_\mathcal{R}$ that optimizes for information gain must \textit{maximize} $\mathcal{T}_{conv}$ and \textit{minimize} $\mathcal{T}_{dist}$. Intuitively, we want the human's action $a$ to be likely for a specific choice of $\theta$ (increasing $\mathcal{T}_{conv}$), but not likely for every possible $\theta$ (decreasing $\mathcal{T}_{dist}$). Return to our motivating example and imagine that the hidden phone is on the left side of the room. For this $\theta$, the interface should select signals $x$ that always cause the human to walk left. However, if $\theta$ changes (i.e., the phone is now on the right side) the robot's signals should not cause the human to keep walking left (and take the same action $a$). Instead, different human actions $a$ should be likely for different choices of interface information $\theta$.

\subsection{Learning to Maximize Information Gain} \label{sec:M3}

With Equations~(\ref{eq:M5})-(\ref{eq:M7}) we now have a formula for information gain in terms of the human and interface policies. Ideally we would optimize over these equations to find the interface policy $\pi_\mathcal{R}$ that maximizes conditional information gain. Unfortunately, this is not possible for two reasons: (a) we do not know the human's current policy $\pi_\mathcal{H}$ and (b) it is intractable to evaluate information gain in continuous $\mathcal{S}$, $\mathcal{A}$, $\mathcal{X}$, and $\Theta$ spaces \cite{belghazi2018mutual, poole2019variational, song2019understanding}. Instead of directly computing the information gain, we here introduce \textbf{L}earning \textbf{I}nterfaces to \textbf{M}aximize \textbf{I}nformation \textbf{T}ransfer (\textbf{LIMIT}). LIMIT is a real-time, personalized \textit{learning} approach that closely mirrors the structure of Equations~(\ref{eq:M5})-(\ref{eq:M7}). As LIMIT gathers data alongside the human operator, it continually learns and updates the interface policy $\pi_\mathcal{R}$ to correlate the human's actions and the hidden information. We emphasize that LIMIT does not have access to the human's task or reward function: this task-agnostic approach learns policies to optimize a \textit{proxy} of conditional information gain.

\p{Models} LIMIT consists of three neural networks. We introduce the first two networks here: let $\mathcal{H}_\phi$ be a model of the human's policy with weights $\phi$, and let $\mathcal{R}_\psi$ be the interface's learned policy with weights $\psi$. The structure of these models corresponds to \eq{P2} and \eq{P3}:
\begin{equation} \label{eq:M8}
    \mathcal{H}_{\phi} : \mathcal{S} \times \mathcal{X} \rightarrow \mathcal{A}, \quad \quad \mathcal{R}_{\psi} : \mathcal{S} \times \Theta \rightarrow \mathcal{X}
\end{equation}
so that $\mathcal{H}_{\phi}$ maps states and signals to actions and $\mathcal{R}_{\psi}$ maps states and hidden information to signals\footnote{When using LIMIT, the interface policy $\pi_\mathcal{R}$ is the learned model $\mathcal{R}_{\psi}$.}. It should be noted that $\mathcal{H}_\phi$ is not the human's actual policy (which the robotic interface never knows). However, while the interface does not observe the human's policy $\pi_\mathcal{H}$ or even their current task, the interface does have access to data from previous interactions. Let $\mathcal{D} = \{(s, x, a, \theta^0), \ldots (s^N, x^N, a^N, \theta^N)\}$ be the dataset of observed states, signals, actions, and hidden information across all previous interactions.

\smallskip

Below we introduce the two loss functions used to train $\mathcal{H}_{\phi}$ and $\mathcal{R}_{\psi}$ on dataset $\mathcal{D}$. These loss functions are analogous to the terms $\mathcal{T}_{conv}$ and $\mathcal{T}_{dist}$ from \eq{M7}.

\p{Convey} Remember from Section~\ref{sec:M2} that interfaces which optimize information gain will maximize the \textit{convey} term in \eq{M5}. Intuitively, $\mathcal{T}_{conv}$ expresses the probability of the human's observed action $a$ given $s$ and $\theta$. We here introduce a loss term analogous to \eq{M5} that specifically applies to our deterministic human and robot models:
\begin{equation} \label{eq:M9}
    \mathcal{L}_{conv}(\phi, \psi) = \sum_{(s, a, \theta) \in \mathcal{D}} \Big\|a - \mathcal{H}_{\phi}\big(s, \mathcal{R}_{\psi}(s, \theta)\big)\Big\|^2
\end{equation}
For any $(s,a,\theta)$ tuple in our dataset, \eq{M9} asserts that the human model and interface policy should map $s$ and $\theta$ to an action that closely matches the human's actual behavior $a$. Put another way, $\mathcal{R}_{\psi}(s, \theta)$ should output a signal $x$ that causes our human model to take the observed action $a$. When comparing \eq{M9} and \eq{M5} we recognize that maximizing $\mathcal{T}_{conv}$ and minimizing $\mathcal{L}_{conv}$ accomplish similar things: both assert that hidden information $\theta$ and action $a$ should be correlated at state $s$, so that if the system observes $(s,\theta)$ the human always outputs $a$.

\p{Distinguish} We next develop a proxy loss function for the \textit{distinguish} term in \eq{M6}. $\mathcal{T}_{dist}$ implies that an optimal interface will result in a one-to-one mapping between hidden information and human actions. Another way to put this is --- given the states and actions output by our models --- the system should be able to accurately infer $\theta$. For instance, during one interaction our LED interface may display signals that guide the human to the left of the room. Based on this sequence of states and human actions we should be able to infer the unique $\theta$ that the interface had in mind (i.e., the phone is on the left of the room). Let $s$ and $\theta$ be sampled from $\mathcal{D}$, and let $\tau(s, \theta) = \big((s, a), (s', a'), \ldots \big)$ be a sequence of $k$ counterfactual states and actions starting at $s$:
\begin{equation} \label{eq:M10}
    a = \mathcal{H}_{\phi}\big(s, \mathcal{R}_{\psi}(s, \theta)\big), \quad\quad s' = f(s, a) 
\end{equation}
Here we use our learned human and interface models and the system dynamics from \eq{P1} to rollout a hypothetical interaction: given that the human starts at $s$ with hidden information $\theta$, sequence $\tau(s, \theta)$ predicts how the system will behave. We then decode this sequence to try and infer $\theta$ from the states and actions:
\begin{equation} \label{eq:M11}
    \mathcal{L}_{dist}(\phi, \psi, \sigma) = \sum_{(s, \theta) \in \mathcal{D}} \Big\|\theta - \Delta_\sigma\big(\tau(s, \theta)\big) \Big\|^2
\end{equation}
where $\Delta_\sigma : \mathcal{S}^k \times \mathcal{A}^k \rightarrow \Theta$ is a decoder model with weights $\sigma$. This decoder is the third and final network within LIMIT. \eq{M11} is minimized when the interface policy $\mathcal{R}_{\psi}(s, \theta)$ outputs signals $x$ that cause the human model $\mathcal{H}_{\phi}$ to take different actions for different $\theta$, enabling the decoder $\Delta_\sigma$ to successfully identify the hidden information $\theta$ behind these actions. Our loss function $\mathcal{L}_{dist}$ is therefore analogous to $\mathcal{T}_{dist}$: minimizing both terms encourages a one-to-one mapping between hidden information and human actions.

\begin{algorithm}[t]
    \setstretch{1.0}
    \caption{LIMIT: Learning Interfaces to Maximize Information Transfer}
    \label{alg:limit}
    \begin{algorithmic}[1]
    \State Initialize model weights $\phi$, $\psi$, and $\sigma$
    \State Initialize dataset $\mathcal{D} \gets \{  \}$
    \For{each interaction}
        \State $\theta \sim P(\theta)$ \Comment{Interface observes hidden information}
        \For{timestep $t = 0 \ldots T$}
        \If{$length(\mathcal{D}) \geq batch\_size$}
        \State Sample batch of recent $(s, a, \theta) \in \mathcal{D}$
        \State Train $\mathcal{H}_{\phi}$, $\mathcal{R}_{\psi}$, and $\Delta_\sigma$ to minimize $\mathcal{L}$
        \EndIf
        \State $s^t \gets $ measured system state
        \State $x^t \gets \mathcal{R}_{\psi}(s^t, \theta)$ \Comment Display signal $x^t$ to human
        \State $a^t \gets $ human action
        \State $\mathcal{D} \gets (s^t, x^t, a^t, \theta^t)$ \Comment Append data to dataset
        \State $s^{t+1} \gets f(s^t, a^t)$ \Comment Transition to next state
        \EndFor
    \EndFor
    \end{algorithmic}
\end{algorithm}

\p{Loss Function} We finally combine \eq{M9} and \eq{M11} to generate the loss function for training LIMIT:
\begin{equation} \label{eq:M12}
    \mathcal{L}(\phi, \psi, \sigma) = \mathcal{L}_{conv}(\phi, \psi) + \mathcal{L}_{dist}(\phi, \psi, \sigma)
\end{equation}
This loss function is used to update the human model $\mathcal{H}_{\phi}$, the interface policy $\mathcal{R}_{\psi}$, and the decoder $\Delta_\sigma$. Note that this loss function is over \textit{continuous space}, not over discrete space like the relation presented in Equation (\ref{eq:M7}). The human model and decoder are purely for training purposes. During interaction the LIMIT interface selects signals $x$ according to the learned model $\mathcal{R}_{\psi}$, and then displays these signals to the actual human operator. Because the model structure and loss function of LIMIT closely mirror \eq{M7}, our proposed LIMIT approach learns to maximize a real-time, tractable proxy of conditional information gain. See Algorithm~\ref{alg:limit} for an outline of LIMIT. To download an implementation of LIMIT, see our repository: \url{https://github.com/VT-Collab/LIMIT-learning-interfaces}. The code in this respository corresponds to the $2$D simulations from Section~\ref{sec:sims}.

As an aside, we recognize that recent works also attempt to estimate information gain \cite{belghazi2018mutual, poole2019variational, song2019understanding}. However, these learning approaches are not applicable to our problem setting because (a) they require offline training data and (b) they maintain a static estimate of information gain. LIMIT learns online, from the current user, and co-adapts alongside the human to maximize a \textit{proxy} of conditional information gain.

\subsection{Accounting for Human Co-Adaptation} \label{sec:M4}

So far we have focused on the interface's perspective, and learned an interface policy that maximizes information gain. Importantly, this interface does not know the human's task or reward function: the interface is learning to correlate humans actions with hidden information, and not necessarily to perform the task correctly. Consider our running example where the human is looking for their missing phone. The interface could learn to turn on the \textit{blue light} when the phone is on the \textit{right} side of the room and the \textit{red light} when the phone is on the \textit{left} side. But the human initially interprets this feedback with the \textit{opposite} mapping: perhaps the human goes left for the blue light and right for the red light. From the interface's perspective this is interpretable behavior that maximizes information gain (i.e., human actions are correlated with $\theta$). But from the human's perspective this is the exact opposite of what we wanted: instead of maximizing reward, the human is guided away from their phone!

We recognize that humans are not static agents; over time, the human will inevitably adapt to the interface. At the end of each interaction the human observes their reward $R(\xi, \theta)$. By reasoning over this reward and the previous signals and actions, the human may shift their policy $\pi_\mathcal{H}$ to improve performance \cite{nikolaidis2017human, parekh2022learning, ikemoto2012physical, van2021becoming}. Returning to our example, once the human realizes that they are going away from the phone using the initial $\pi_\mathcal{H}$, they may switch their interpretation so that they correctly go right for the blue light and left for the red light. Of course, this adaptation is not a one-way street; as the human adapts to the interface, LIMIT should also adapt to the human's changing policy. We explicitly encourage personalization by biasing the system's learning towards \textit{recent} human data. When sampling $(s,a,\theta)$ tuples  in Algorithm~\ref{alg:limit}, we set the probability of sampling the most recent data $(s^N, a^N, \theta^N)$ as exponentially more likely than sampling the first datapoint $(s^0, a^0, \theta^0)$. Overall, LIMIT learns an interface policy that transfers hidden information to the human; the human is then responsible for taking advantage of this information and maximize task reward. We will test how LIMIT adapts to the human --- and how the human co-adapts to the learning interface --- through our simulations and user studies.

\section{Simulations} \label{sec:sims}

We first compare our LIMIT algorithm to naive alternatives and a state-of-the-art baseline across controlled simulations. Within these simulations the interface knows the 
{hidden information} 
$\theta$ (e.g., the location of the human's phone), and the human is 
{trying to maximize their task reward $R(\xi, \theta)$ (e.g., reach their phone by the end of each interaction).}
The interface's signal is a vector $x$, and the simulated human interprets this signal to select their own action $a$. Importantly, our simulated humans are adaptive agents; they change how they interpret the signals between interactions based on their past experiences and observed rewards. The interface must learn to personalize alongside these shifting humans and accurately convey the hidden information.

\p{Interface Algorithms} We compare five different methods for selecting the feedback signals:
\begin{itemize}
    \item \textbf{Naive.} The interface multiplies the vector $(s, \theta)$ by a randomized matrix to get signal $x$. The matrix elements are uniformly randomly sampled between $[-1, +1]$.
    \item \textbf{Bayes \cite{reddy2022first}.} The interface uses a matrix to map $(s, \theta)$ to $x$. At the end of each interaction the interface observes the task reward $R(\xi, \theta)$. The elements of the matrix are updated to maximize this reward using Bayesian optimization \cite{bayes}.
    \item \textbf{Convey.} An ablation of our approach where the interface policy is only trained with loss $\mathcal{L}_{conv}$ in \eq{M9}.
    \item \textbf{Distinguish.} An ablation of our approach where the interface policy is only trained with loss $\mathcal{L}_{dist}$ in \eq{M11}.
    \item \textbf{LIMIT.} Our proposed approach from Algorithm~\ref{alg:limit}. For the implementation of LIMIT used in these simulations see \url{https://github.com/VT-Collab/LIMIT-learning-interfaces}
\end{itemize}
We note that the \textbf{Bayes} method adopted from \cite{reddy2022first} has access to the human's reward function, and uses this reward function when designing the feedback signals. By contrast, within our problem setting we \textit{do not assume} any knowledge of $R(\xi, \theta)$. So while we believe that \textbf{Bayes} is the closest existing alternative to \textbf{LIMIT}, it is important to remember that \textbf{Bayes} knows the human's reward function while \textbf{LIMIT} is task-agnostic.

\p{Simulated Humans} In Sections~\ref{sec:sims-1d}--\ref{sec:sims-3d} we pair the interface with two types of simulated humans: rotate and align. Both types of simulated humans take actions based on signal $x$, where $x$ is a vector with elements bounded between $[-1, +1]$.
\begin{itemize}
    \item \textbf{Rotate.} This simulated human \textit{rotates} $x$ to get action $a$. In our $1$D environment the human multiplies $x$ by $+1$ or $-1$ (to change the sign). In the $2$D environment the human multiplies $x$ by an $SO(2)$ rotation matrix.
    \item \textbf{Align.} This simulated human \textit{rotates and scales} $x$ to get action $a$. The rotation is the same as \textbf{Rotate}. For scaling, the human multiplies $x$ by a value between $[-1, +1]$.
\end{itemize}
Both types of simulated humans update their rotation (and scaling) at the end of each interaction to \textit{co-adapt} to the interface. In Sections~\ref{sec:sims-1d}--\ref{sec:sims-3d} the human is attempting to reach their missing phone: the reward function is the negative distance between the human's final position and the phone position: $R(\xi, \theta) = -\| s^t - \theta\|^2$. In order to co-adapt to the interface the human randomly samples $N$ recent interactions and finds the rotation matrix (and scalar) that would have maximized reward over those $N$ interactions. Put another way, the human co-adapts so that their mapping from signals to actions would have increased their task reward over recent interactions.

\begin{figure*}[t]
    \begin{center}
    \includegraphics[width=1.0\columnwidth]{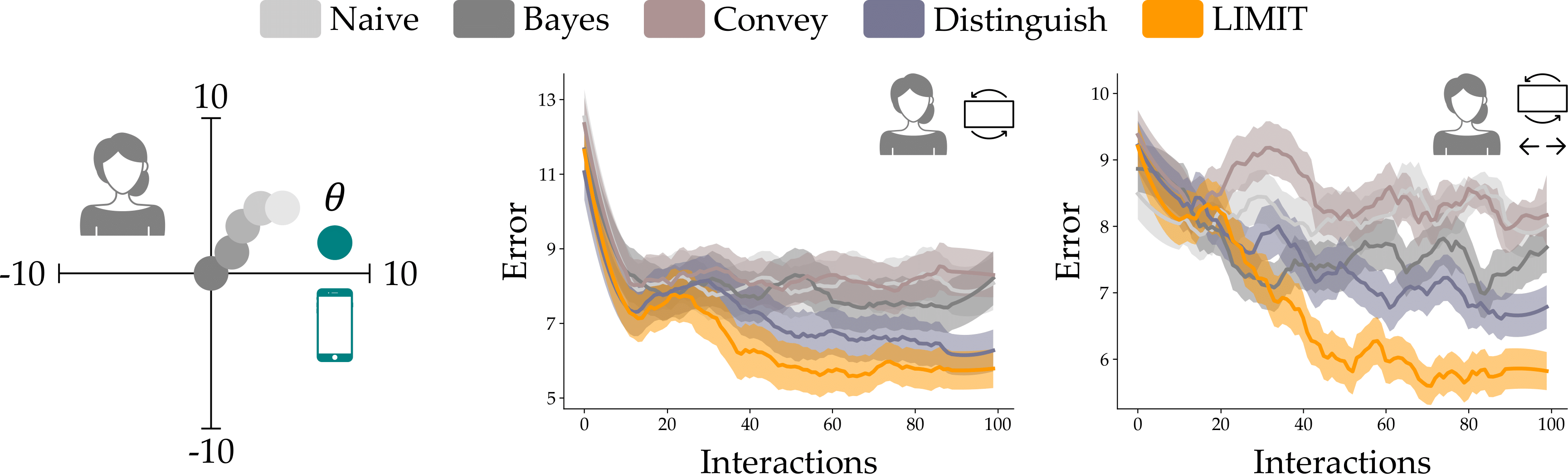}
		\caption{Simulation results in $2$D environment. (Left) The human observes vector $x$ and tries to reach hidden location $\theta$. (Middle) Results with \textbf{Rotate} human: differences here are not statistically significant. (Right) Interface paired with an \textbf{Align} human. Here interface type has a significant effect on error ($F(4,196)=9.0$, $p<.001$), and humans using \textbf{LIMIT} have less error by the final interaction than humans using \textbf{Naive}, \textbf{Bayes}, or \textbf{Convey} ($p<.001$).}
		\label{fig:sim2}
	\end{center}
\end{figure*}

\p{Environments} Our simulated environments are shown in \fig{sim1} and \fig{sim2}. The $1$D environment is a number line and the $2$D environment is an $x$-$y$ plane. At the start of each interaction the human begins at the origin and the interface samples a random phone position $\theta$. The human and interface interact for $10$ timesteps: during each timestep the interface displays signal $x$, the human takes action $a$, and the state transitions according to $s^{t+1} = s^t + a^t$. At the end of the interaction we measure the distance $\| s^t - \theta\|$ between the human and their missing phone. We emphasize that $\mathcal{S}$, $\mathcal{A}$, $\mathcal{X}$, and $\Theta$ are continuous spaces, and the hidden information $\theta$ is known only by the interface.

\subsection{Single-DoF Environment}\label{sec:sims-1d}

\begin{figure*}[t]
    \begin{center}
    \includegraphics[width=1.0\columnwidth]{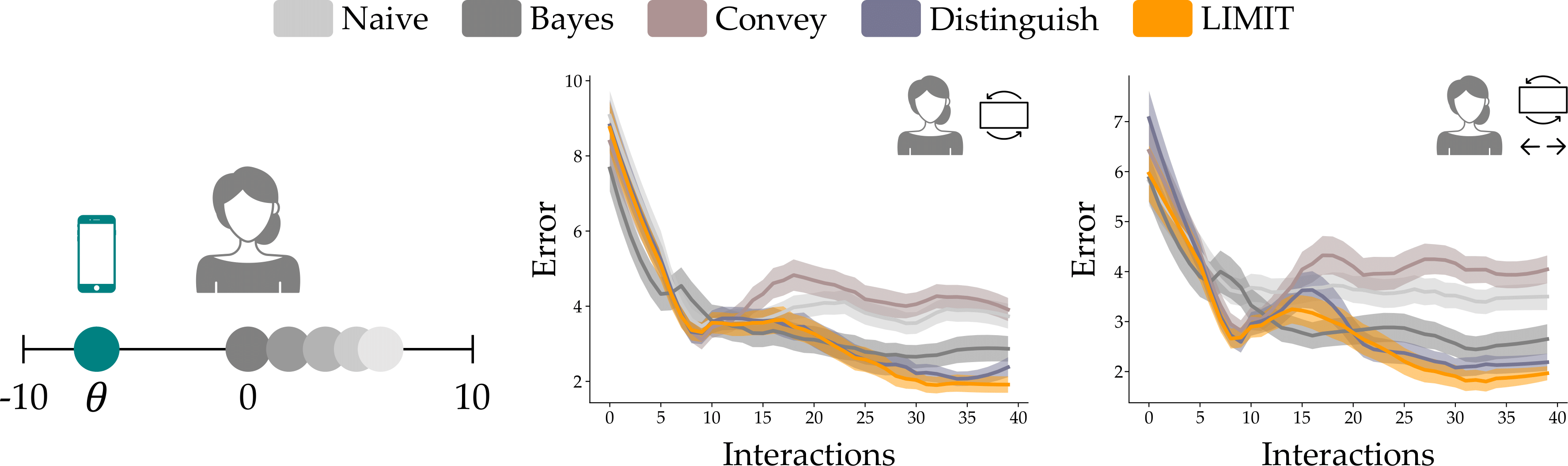}
		\caption{Simulation results in $1$D environment. (Left) Error is the distance between the human's final position and the phone location $\theta$. (Middle) Interfaces paired with the \textbf{Rotate} human. A repeated measures ANOVA reveals that the interface type had a significant effect on error ($F(4, 396) = 9.2$, $p < .001$), with \textbf{LIMIT} resulting less error than the alternatives ($p<.05$). (Right) Interfaces paired with the \textbf{Align} human. The interface algorithm affects error ($F(4, 396) = 16.8, p<.001$); pairwise comparisons show that \textbf{LIMIT} leads to less error than all alternatives besides \textbf{Distinguish} ($p<.05$).}
		\label{fig:sim1}
	\end{center}
\end{figure*}

We start with the $1$D environment. Here the state, signal, action, and hidden information are all scalars. The simulated human and interface collaborate across $40$ interactions: at the start of each interaction $\theta$ is sampled uniformly at random from $[-10, +10]$, and we measure the \textit{error} between the final state and $\theta$. The averaged results across $100$ simulations are reported in \fig{sim1}. For humans that co-adapt using \textbf{Rotate} or \textbf{Align} we find that \textbf{LIMIT} leads to the lowest average error (i.e., \textbf{LIMIT} users most accurately reach their phone).

\subsection{Two-DoF Environment}\label{sec:sims-2d}

We next perform the same experiment in a $2$D environment where the state, signal, action, and $\theta$ are all two-dimensional and continuous vectors. The simulated human and interface work together across $100$ interactions, and the human's position is reset to the origin $(0, 0)$ at the start of each interaction. We display the averaged results for $50$ simulations in \fig{sim2}. Decreasing \textit{error} indicates that --- for each interface algorithm --- the distance between the human's final state and the hidden $\theta$ decreases over interactions. However, we again find that \textbf{LIMIT} has the lowest mean error with our \textbf{Rotate} and \textbf{Align} humans.

\subsection{Mismatch between Signals and Information} \label{sec:sims-3d}

\begin{figure*}[t]
    \begin{center}
    \includegraphics[width=1\columnwidth]{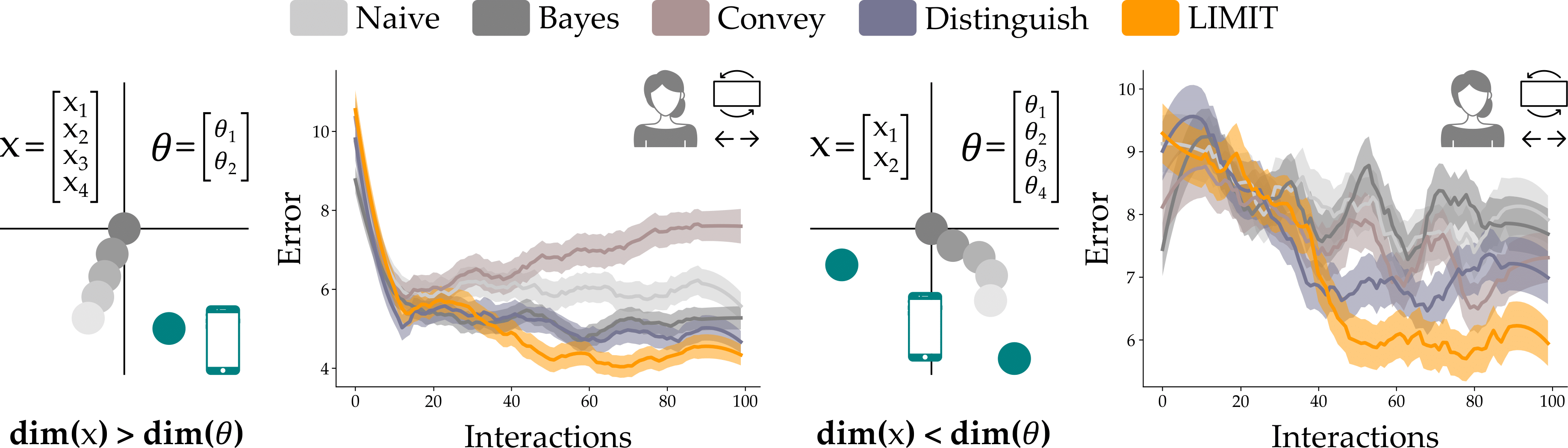}
		\caption{Simulation results in the $2$D environment when the signal $x$ and hidden information $\theta$ have different dimensions. (Left) The interface signal is $4$-dimensional, but only two dimensions are necessary to convey position $\theta$. Interface type has a significant effect on error ($F(4,196)=10.2$, $p<.001$) and \textbf{LIMIT} results in lower final error than either \textbf{Naive} or \textbf{Convey} ($p<.05$). (Right) Now the hidden information is $4$ dimensional, and the interface must embed this $\theta$ to a lower-dimensional signal $x$ (e.g., the interface is trying to convey two phone locations). Humans reach different errors with different methods ($F(4,196)=4.6$, $p<.001$), but \textbf{LIMIT} yields less error than all baselines ($p<.05$).}
		\label{fig:sim3}
	\end{center}
\end{figure*}

In our next simulation we focus on the $2$D environment and the \textbf{Align} human. We vary the dimensions of $x$ and $\theta$ to test scenarios where the interface has additional feedback channels, $dim(x) > dim(\theta)$, or where the hidden information is more complex than the interface, $dim(x) < dim(\theta)$. On the left of \fig{sim3} the signal $x$ is a $4$-dimensional vector: because $\theta$ here is only an $(x,y)$ position, the robot must learn how to harness two additional feedback dimensions. The second scenario is shown on the right side of \fig{sim3}. Here $\theta$ is a $4$D vector specifying the position of two hidden phones: the interface must learn to embed this higher-dimension hidden information into a $2$-dimensional signal $x$. Within these controlled environments with simulated users, we again observe that \textbf{LIMIT} outperforms the baselines.

{

\subsection{LIMIT in more Complex Tasks}\label{adx:complex}

To test the effectiveness of LIMIT in more complex tasks, we present two additional simulations. 
These simulations vary in complexity along two axes: the type of information the interface needs to convey to the human, and the dimensionality of the problem setting.
We start with an autonomous driving scenario, where the interface attempts to convey the policy of an autonomous car to a nearby human driver. We then end this section by returning to the phone example, but now in a $10$-dimensional state-action space.
Note that --- in these more complex settings --- we model the human as a multi-layer perceptrons with one hidden layer.
At the end of each interaction, this simulated human co-adapts to the interface by updating the weights of its model to maximize the measured reward.
We also conduct supplementary simulations to explore the adaptation rate and human variability in the Appendix (Section~\ref{adx:sim}).

\p{Autonomous Driving Task} In this simulation the human agent is driving along a two-lane one-way highway (see \fig{adx-road}). Ahead of the human is an autonomous vehicle that the human would like to avoid; the autonomous vehicle may be in either lane of the highway. The autonomous vehicle attempting to signal its \textit{policy} (i.e., how it will change lanes). There are four discrete policies that the autonomous vehicle could be using to drive along the highway:
\begin{enumerate}
    \item The robot will always stay in the right lane ($\theta_1$)
    \item The robot will always stay in the left lane ($\theta_2$)
    \item The robot will merge into the human's lane ($\theta_3$)
    \item The robot will merge into the opposite lane of the human ($\theta_4$)
\end{enumerate}
Each interaction lasts five timesteps, after which the human car and autonomous car are reset. The autonomous car's policy is randomly sampled at the start of the interaction: this policy is the autonomous car's hidden information $\theta$. The human agent is penalized when it collides with the autonomous car, so it should learn to infer $\theta$ from the signals produced by the interface. To test the effectiveness of LIMIT, we compared the performance of LIMIT to our \textbf{Naive} baseline and the state-of-the-art \textbf{Bayes} baseline. Figure \ref{fig:adx-road} (Left) shows the average collision rate per interaction over time; LIMIT outperforms the baselines, approaching a collision rate of $0$. This result suggests that LIMIT can be successfully applied in scenarios where the interface needs to convey more complex information to the human (i.e., the policy of another agent).

\begin{figure*}[t]
    \begin{center}
    \includegraphics[width=1.0\columnwidth]{Figures/complex_tasks.png}
		\caption{(Left) Simulation results from the autonomous driving task. The autonomous car has four different driving policies, and must signal its current policy to the human in order to help the human driver avoid a collision. Interfaces generated by \textbf{LIMIT} result in a lower collision rate than either \textbf{Naive} or \textbf{Bayes} ($p < 0.001$). (Right) Results from a $10$-dimensional environment. This simulation extends Section~\ref{sec:sims-2d} to a high-dimensional setting where the states, actions, signals, and hidden information are all $10$-dimensional vectors in a continuous space. Despite this increase in dimension, the interfaces generated by \textbf{LIMIT} still result in a lower error at the end of an interaction than those generated by \textbf{Naive} or \textbf{Bayes} ($p < 0.001$).}
		\label{fig:adx-road}
	\end{center}
\vspace{-0.5em}
\end{figure*}

\p{10-DoF Environment} In this simulation, we replicate the experiments from Sections \ref{sec:sims-1d} and \ref{sec:sims-2d}, but now increase the dimension of the state, action, and hidden information so that each component is a $10$-dimensional continuous space. The interface knows the goal position $\theta$, and attempts to convey this hidden information through $10$-dimensional signals. The simulated human attempts to reach the hidden goal position over $10$ timesteps based on the signals from the interface. Like before, we compare the performance of LIMIT against the \textbf{Naive} and \textbf{Bayes} algorithms. Figure \ref{fig:adx-road} (Right) shows that LIMIT outperforms the baselines --- despite the high-dimensionality of the problem setting --- while the performance of \textbf{Bayes} and \textbf{Naive} follow random error. This result indicates that LIMIT can be extended to higher-dimensional problem settings.

}

\section{Can Users Understand Learned Interfaces?} \label{sec:amt}

Our simulations from Section~\ref{sec:sims} suggest LIMIT learns interface policies that better convey hidden information than the alternatives. However, these tests were run with simulated humans in controlled environments. How does LIMIT fare with actual users? To evaluate the real-world performance, we first conducted an online user study via Google Forms where participants observed colored signals and attempted to guess the $2$D position of their missing phone. Here the interfaces were trained offline --- using synthetic human data --- and we measured whether participants (a) found the learned mappings intuitive and (b) adapted to the interface policy. Our results across $37$ online participants indicate that LIMIT interfaces are more interpretable than a randomized baseline.

\p{Independent Variables} We compared \textbf{LIMIT} to a \textbf{Naive} baseline. During each interaction the human started at state $s = (0, 0)$, observed a signal $x$, and then clicked \textit{once} to indicate where they thought the phone was hidden. The interface had access to the phone's hidden location $\theta$. \textbf{Naive} multiplied the vector $\theta$ by a randomized $2 \times 2$ matrix to get signal $x$. \textbf{LIMIT} was pre-trained offline using the procedure from Section~\ref{sec:M2}. After training, we recorded the signals produced by \textbf{Naive} and \textbf{LIMIT} for nine different $\theta$ positions.

\begin{figure}[t]
    \begin{center}
    \includegraphics[width=0.6\columnwidth]{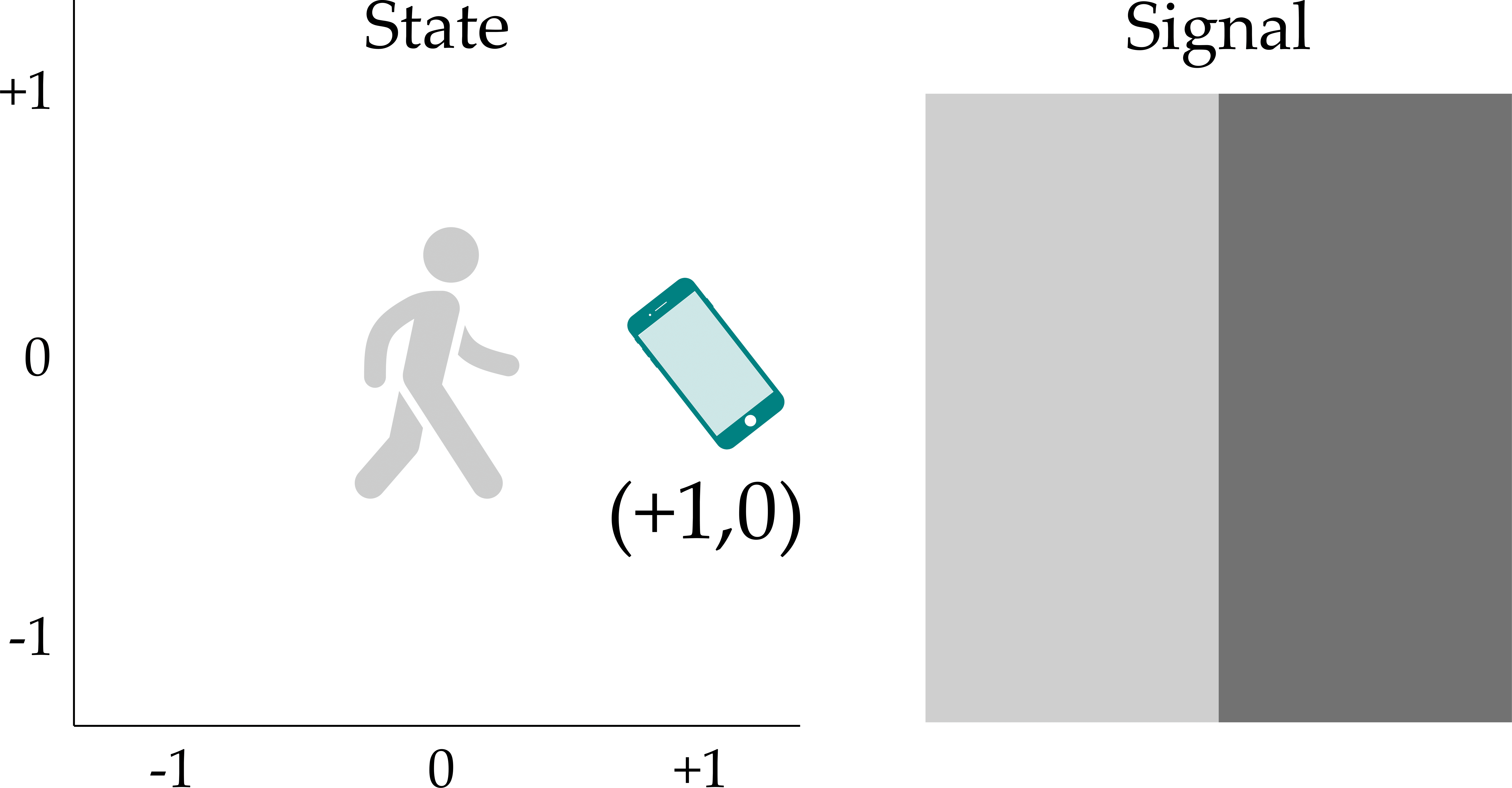}
		\caption{Example of a state and signal displayed to online study participants in Section~\ref{sec:amt}. At first the phone's location $\theta$ (in blue) was \textit{not} shown: participants just saw the human and the signal, and tried to guess the phone's location based on the color of the two bars. For instance, perhaps the color of the first bar corresponded to the $x$-axis and the color of the second bar corresponded to the $y$-axis. We then revealed $\theta$ and moved on to a new state and signal.}
		\label{fig:amt1}
	\end{center}
\end{figure}

\p{Experimental Setup} At the start of each interaction the online participants were shown a picture of their current position $s$ and two colored bars for signal $x$ (see \fig{amt1}). After observing the state and signal, the participants selected the $x$-$y$ coordinates that they thought the interface was trying to convey. We then displayed the phone's actual location $\theta$. Because we revealed the hidden information $\theta$ after each interaction, users could adapt to the interface over time. 

Each participant completed nine interactions with the \textbf{Naive} approach and nine interactions with \textbf{LIMIT}. The order of the methods was counterbalanced: half of the participants started with \textbf{Naive} and the other half started with \textbf{LIMIT}.

\p{Dependent Variables} To determine how participants objectively performed with each interface, we measured the distance between their guess $s^1$ and the phone's actual position $\theta$. Specifically, we defined \textit{Error} as $\|s^1 - \theta\|$. To understand how participants subjectively perceived each interface, we administered a 7-point Likert scale survey each time the users completed a method. Questions were organized along three scales: how confident users were that their performance \textit{improved} over time, if they \textit{understood} what the interface was trying to convey, and whether the interface was \textit{intuitive}. The exact items on the survey are listed in Section~\ref{sec:appendix}.

\p{Participants} A total of $38$ adults took part in this online survey. At the start of the survey we asked participants to read the instructions, follow an example interaction, and then answer three qualifying questions to test their understanding. Below we report the results for $37$ participants who correctly answered these questions and finished the survey.

\begin{figure}[t]
    \begin{center}
    \includegraphics[width=0.8\columnwidth]{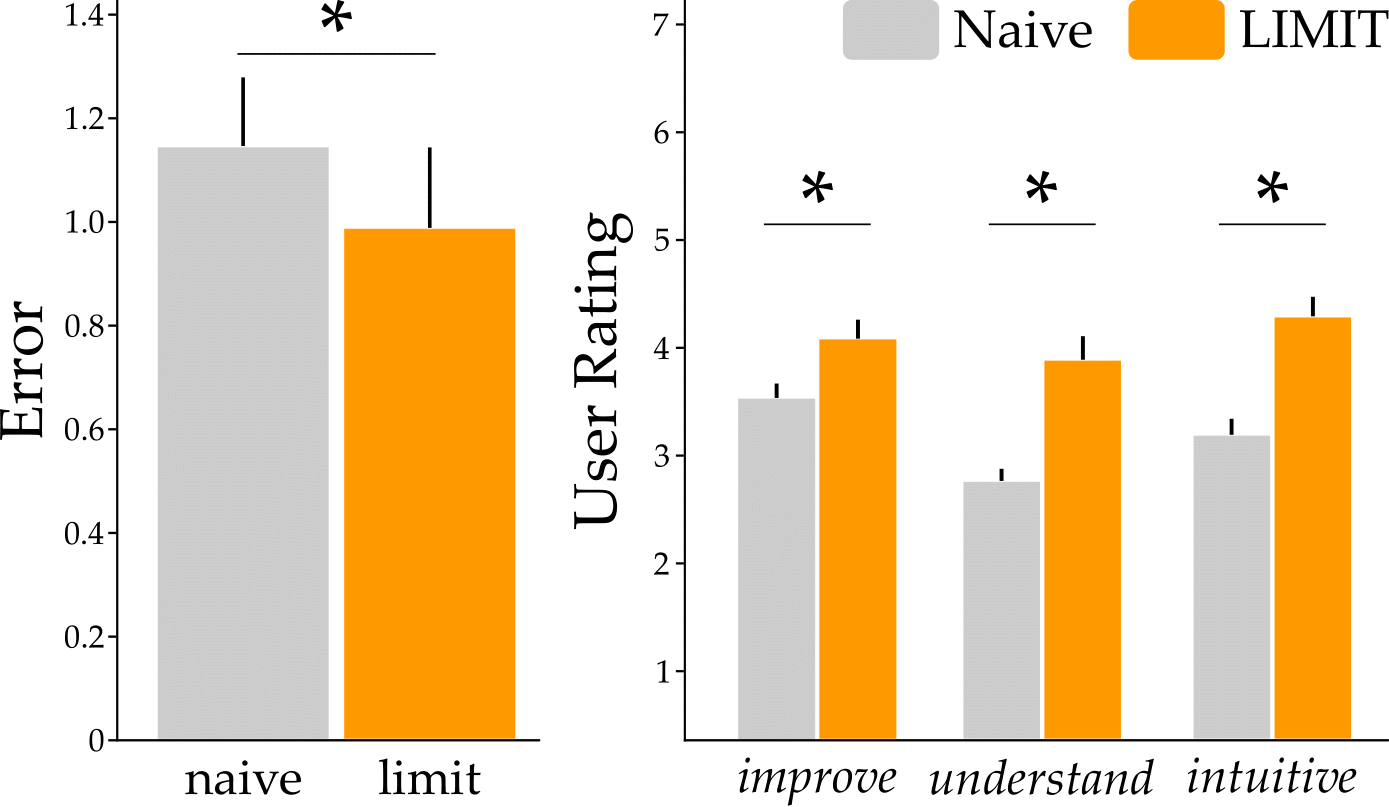}
		\caption{Results from the online user study in Section~\ref{sec:amt}. (Left) Our $37$ participants interpreted the signals in \fig{amt1} to predict the phone's hidden position. The error between their guess and the actual position was lower with \textbf{LIMIT} than with \textbf{Naive}. (Right) With \textbf{LIMIT} participants thought they better improved over time, better understood what the interface was trying to convey, and overall perceived the interface as more intuitive. Error bars show standard error and $*$ denote statistical significance ($p < .05$).}
		\label{fig:amt2}
	\end{center}
\end{figure}

\p{Results} Our results are summarized in \fig{amt2}. We first measured the \textit{error} between the human's predictions with \textbf{Naive} and \textbf{LIMIT} across nine rounds of interaction. Paired $t$-tests revealed that users were significantly more accurate with \textbf{LIMIT} than with \textbf{Naive} ($t(295)=2.17$, $p < .05$). Put another way, when working with \textbf{LIMIT} the online participants were better able to infer the hidden information and select high-reward states. The subjective responses suggested that participants perceived a difference between the methods. To analyze the Likert results we first confirmed that our three multi-item scales (\textit{improve}, \textit{understand}, and \textit{intuitive}) were reliable with a Cronbach's $\alpha > 0.7$. We then grouped each scale into combined scores and performed paired $t$-tests. Overall, our $37$ users thought that they \textit{improved} more with \textbf{LIMIT} ($t(36)=-2.41$, $p<.05$), better \textit{understood} what the \textbf{LIMIT} interface was trying to communicate ($t(36)=-4.31$, $p<.001$), and found the \textbf{LIMIT} interface to be more \textit{intuitive} ($t(36)=-4.16$, $p<.001$).

We emphasize that --- within this online user study --- each interaction only lasted a single timestep, and the pre-trained interface policy did not learn or adapt alongside actual user data. However, our results across $37$ participants are a first step towards confirming that the interfaces learned using \textbf{LIMIT} are interpretable for everyday human users.

\section{User Study} \label{sec:user-study}

Our online study was a first step towards evaluating LIMIT. To understand how LIMIT performs across repeated human-robot interaction with different types of interfaces, we next performed an \textit{in-person} user study. In this study participants completed three separate tasks that each had different interface modalities: \textit{sounds}, \textit{lights}, and \textit{haptics}. The interface knew some hidden information $\theta$ (e.g., the correct joint position for a robot arm), and selected feedback signals to convey $\theta$ to the human. There were no immediately obvious conventions for interpreting this feedback. For instance, the interface played musical notes of varying pitches to indicate where to guide a robot arm: one person might assume higher notes indicate moving the arm to the right, while others might think lower notes denote the same motion. Over multiple timesteps and interactions the interface and participant co-adapted. The interface had to learn how to map $\theta$ to signals, and users had to learn to interpret these signals and complete the task.

\p{Independent Variables} We compared two algorithms for selecting feedback: \textbf{Bayes} and \textbf{LIMIT}. \textbf{Bayes} is a state-of-the-art approach adapted from \cite{reddy2022first} that treats the mapping from signals to rewards as a black box. The interface explores this black box by using Bayesian optimization \cite{bayes} to search for signals that maximize task reward. We note that \textbf{Bayes} knows what task that the human is trying to complete (i.e., in this baseline the interface has access to the human's reward function). As such, \textbf{Bayes} actually has more information than \textbf{LIMIT}, where the interface never knows the human's objective. For \textbf{LIMIT} we used our proposed approach from Algorithm~\ref{alg:limit}. Both methods were pre-trained offline with simulated partners using the procedure from Section~\ref{sec:M2}. 

\begin{figure*}[t]
    \begin{center}
    \includegraphics[width=1\columnwidth]{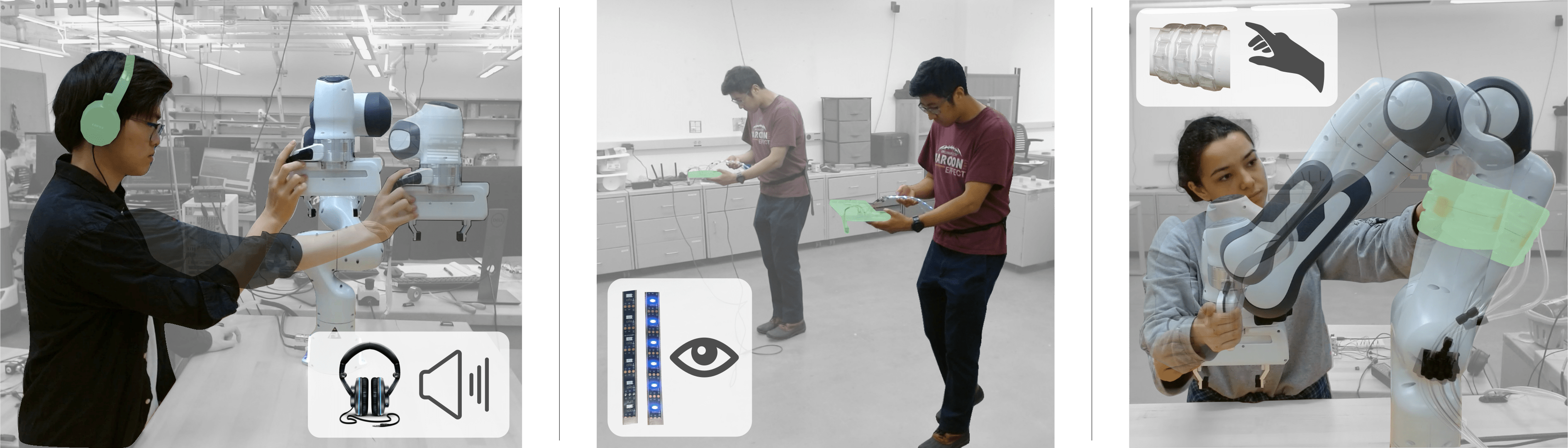}
		\caption{Interfaces and tasks for our in-person study. (Left) In \textit{Sounds} the human moved a robot arm back and forth along a single axis. The interface played a musical note of varying pitches to convey the correct arm position. (Middle) In \textit{Lights} participants walked around an empty space. The interface illuminated two LED strips to indicate the location of their missing phone. (Right) in \textit{Haptics} we wrapped pneumatic bags around a robot arm \cite{valdivia2023wrapping}. Participants needed to guide the robot to the correct height, orientation, and distance from their body: the interface inflated and deflated the haptic bags to convey these features.}
		\label{fig:interfaces}
	\end{center}
\end{figure*}

\p{Interfaces} To demonstrate that our work is not tied to any specific type of interface, we performed tests with interfaces that employed sounds, lights, or haptics (see \fig{interfaces}).
\begin{itemize} [leftmargin=*]
    \item \textbf{Sounds.} At each timestep the system played a musical note (G) through speakers and headphones. The signal $x$ was the $1$-DoF pitch of this musical note: the interface could continuously vary the pitch along two octaves.
    \item \textbf{Lights.} This interface was similar to the online user study in Section~\ref{sec:amt}. Users carried a interface with two strips of LED lights. The $2$-DoF signal $x$ was the brightness of the lights on each strip: at each timestep the interface could change the number and intensity of illuminated LEDs.
    \item \textbf{Haptics.} Here we wrapped three pneumatic bags around a robot arm \cite{valdivia2023wrapping}. Users kinesthetically interacted with these bags as they moved the robot. The $3$-DoF signal $x$ was the pressure of each bag: at every timestep the system could increase or decrease the pressures between $0$ and $3$ PSI.
\end{itemize}
We emphasize that the interfaces were used separately. Users completed tasks with only sounds, lights, or haptics, and did not interact with more than one interface at a time.

\p{Experimental Setup} We divided the study into three tasks, one for each of the interfaces (see \fig{interfaces}). Note that the interface used in the task corresponds to the task name. 

In \textit{Sounds} participants physically interacted with a $7$-DoF Franka Emika robot arm. The robot started each interaction in its home position, and users kinesthetically guided the robot to move its end-effector. The system state $s$ was the position of the end-effector. Users were free to select their final $x$ and $z$ coordinates; here $\theta$ corresponded to the correct position along the $y$-axis. The robot used auditory feedback (the pitch of a musical note) to indicate $\theta$.

\begin{figure*}[t]
    \begin{center}
    \includegraphics[width=0.8\columnwidth]{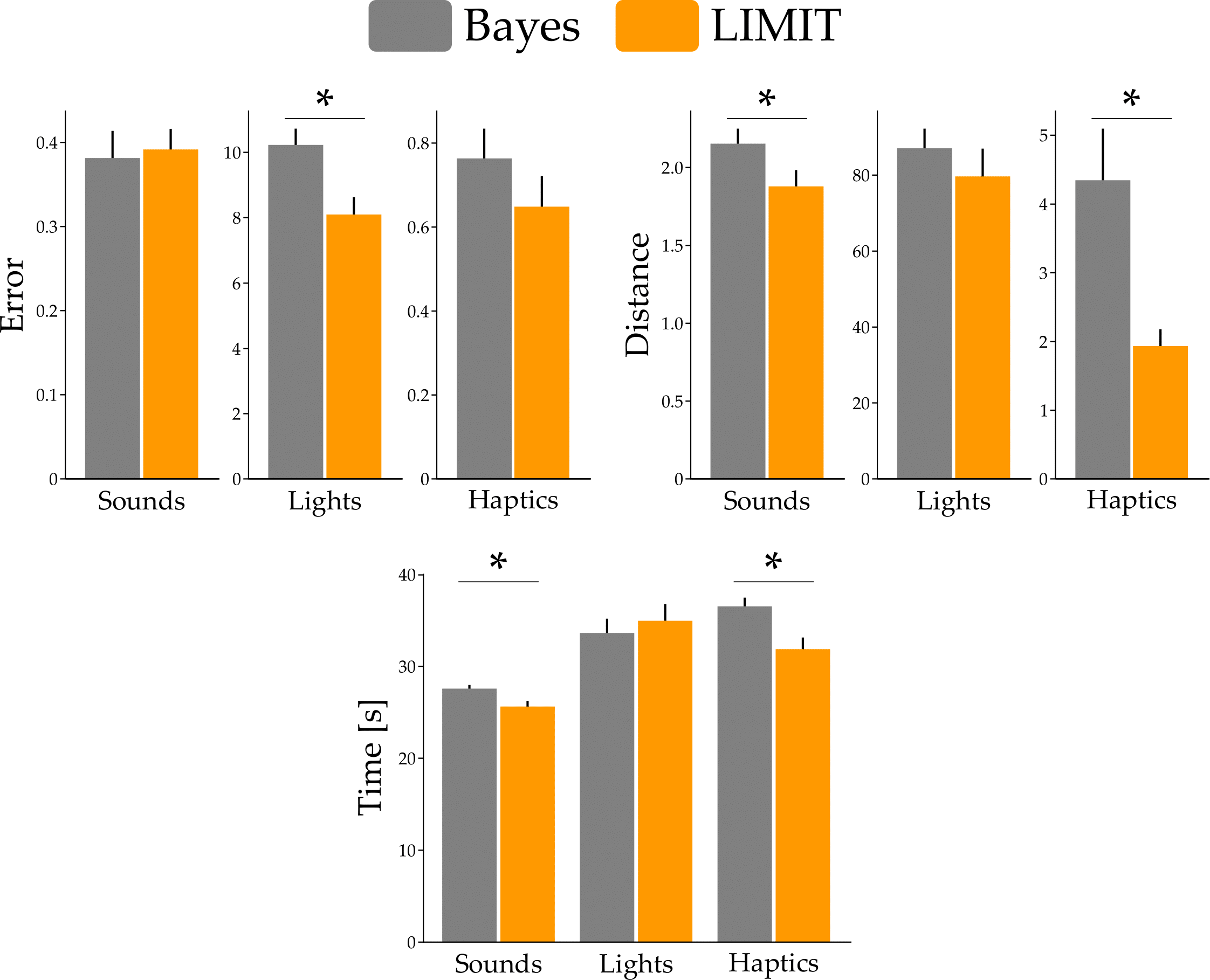}
		\caption{Objective results from our in-person user study. These results are broken down by task; \textit{Sounds} and \textit{Haptics} are performed with a robot arm. (Left) Error between $\theta$ and human's final state $s^T$. (Middle) Distance the human travels during an interaction. For error and distance the units vary between tasks: in \textit{Sounds} the units are meters, in \textit{Lights} the units are feet, and in \textit{Haptics} the units are meters (end-effector position) plus radians (end-effector orientation). (Right) Time taken to complete an interaction. Error bars show standard error and an $*$ denotes statistical significance ($p<.05$).}
		\label{fig:objective}
	\end{center}
\end{figure*}

\begin{figure}[t]
    \begin{center}
    \includegraphics[width=0.6\columnwidth]{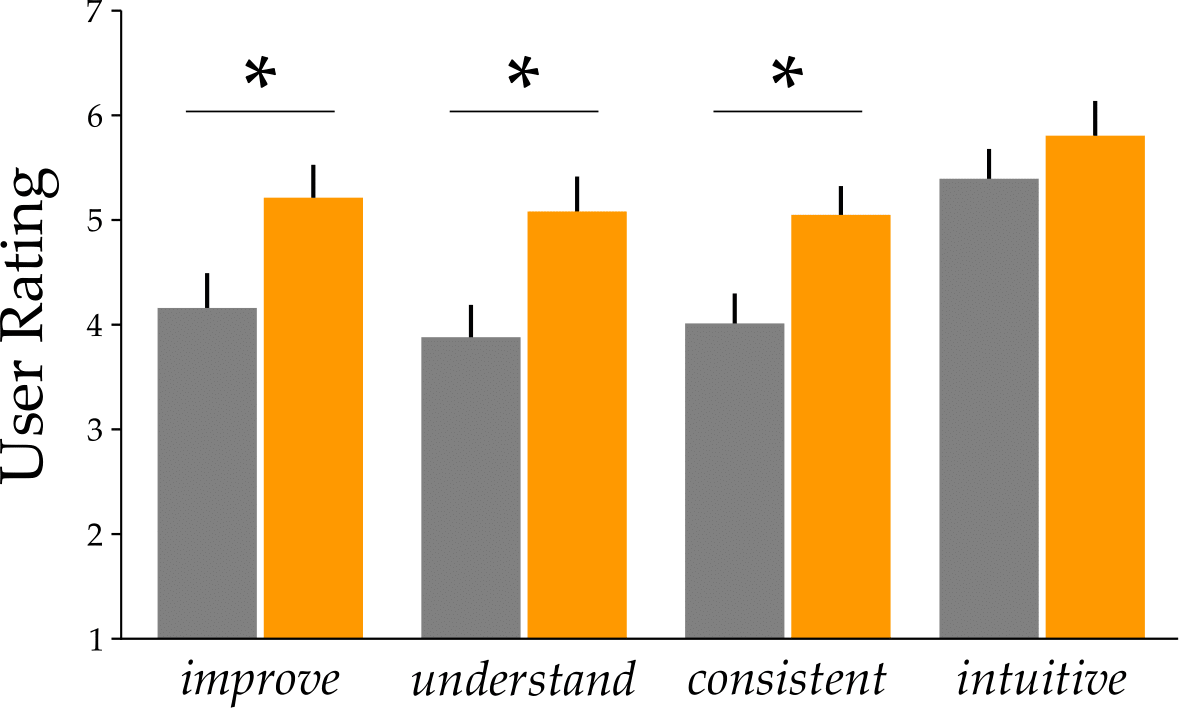}
		\caption{Subjective results from our $7$-point Likert scale survey. Participants thought that they improved more with \textbf{LIMIT}, and that \textbf{LIMIT} signals were more understandable and consistent. Users also scored \textbf{LIMIT} as more intuitive than \textbf{Bayes}, but this was not statistically significant. An $*$ denotes significance ($p<.05$).}
		\label{fig:subjective}
	\end{center}
\end{figure}

\textit{Lights} matches our motivating example of a person walking around a room to find their missing phone. Participants moved in an empty $10$ft by $10$ft space while wearing an HTC Vive position tracker for real-time measurements: the state $s$ was the user's position. Hidden information $\theta$ corresponded to the correct $x$-$y$ position of their missing phone. As users walked they carried our lights interface with two LED strips. Once the user was confident that they had reached the correct position (i.e., they thought they had found their phone), they informed the proctor and the interaction ended.

For \textit{Haptics} participants again interacted with the $7$-DoF Franka Emika robot arm. In this task the robot followed a parameterized trajectory: the robot had a fixed start, but it was up to the user to correct the robot's goal. Here state $s$ was the position and orientation of the end-effector. The robot knew where the endpoint the trajectory should be: hidden information $\theta$ contained the correct height, yaw, and distance from the person. As users physically corrected the robot's trajectory they got haptic feedback from the wrapped display. Note that in \textit{Sounds} $\theta$ and $x$ are $1$-dimensional, in \textit{Lights} they are $2$-dimensional, and in \textit{Haptics} they are $3$-dimensional.

Each task contained multiple interactions. At the start of the interaction we reset the environment (i.e., the robot returned to its home position or the human walked to the center of the room). The interface then sampled a random $\theta \sim P(\cdot)$ from a uniform prior: this $\theta$ was held constant during the interaction but changed between interactions. The interaction lasted multiple timesteps as the human physically guided the robot or walked around the room. At the end of the interaction the system revealed the actual $\theta$ back to human: in \textit{Sounds} and \textit{Haptics} the robot arm moved to the correct pose, and in \textit{Lights} the proctor showed the person the correct place to stand. The \textit{Sounds} and \textit{Lights} tasks included $10$ interactions, and the \textit{Haptics} task had $5$ interactions. Note that the input and output dimensions of each algorithm were changed to accommodate the change in DoF between tasks; all other architectural changes are addressed in Appendix \ref{adx:arch}.

\p{Participants and Procedure} We recruited $11$ participants ($3$ female, ages $25.5 \pm 5.15$) from the Virginia Tech community. Prior to the experiment all participants provided informed written consent consistent with university guidelines (IRB $\#20$-$755$). Two of the eleven participants had never interacted with robots before. None of these in-person users took part in the online study.

Each participant completed all three tasks twice: once with \textbf{Bayes} and once with \textbf{LIMIT}. The order of the tasks was counterbalanced: e.g., some users started with \textit{Sounds} while others started with \textit{Haptics}. The order of the algorithms was also counterbalanced: for each task, half of the users started with \textbf{LIMIT}. Participants were never told which algorithm they were working with during the experiment, and did not know which algorithm was our approach.

\p{Dependent Measures -- Objective} We measured the states, signals, and actions during each timestep. Recall that in every task the human was trying to reach $\theta$ (e.g., the correct robot position). We therefore recorded the \textit{Error} between the system's final state $s^T$ and the desired position $\theta$, so that $Error = \| s^T - \theta\|$. To assess how the human behaved within the task, we also measured the total \textit{Distance} they traveled during an interaction: $Distance = \sum_{t = 1}^T \|s^t - s^{t-1}\|$. Here lower distances indicate that the human understood the interface and went directly to the goal, while higher distances suggest the human often backtracked or changed directions. Finally, we measured the \textit{Time} it took to complete the task.

\p{Dependent Measures -- Subjective} After each task and algorithm participants completed a $7$-point Likert scale survey. This survey was designed to measure the user's subjective perception of the interface along four multi-item scales. We asked users if they felt like their performance \textit{improved} over time, if they could \textit{understand} what the interface was trying to communicate, if the signals seemed \textit{consistent}, and whether they thought they would continue to improve if they kept working with this interface (\textit{intuitive}).

\p{Hypotheses} We had two hypothesis for the user study:
\begin{quote}
\p{H1} \textit{Users will have less error and complete the task more efficiently with \textbf{LIMIT}.}
\end{quote}
\begin{quote}
\p{H2} \textit{Users will subjectively prefer interfaces that use \textbf{LIMIT} to learn feedback signals.}
\end{quote}

\p{Results} The objective results are summarized in \fig{objective}, and the subjective results are displayed in \fig{subjective}. Please also see videos of the user study here: \url{https://youtu.be/IvQ3TM1_2fA}

To explore hypothesis \textbf{H1} we analyzed the error, distance travelled, and time taken when getting feedback from \textbf{Bayes} or \textbf{LIMIT}. For each of these metrics lower was better: an effective interface should help the human complete the task correctly and efficiently. Paired $t$-tests revealed that participants reached significantly lower error when working with \textbf{LIMIT} in the \textit{Lights} task ($t(109)=3.02$, $p<.05$). Interestingly, the difference in error was not significant for either \textit{Sounds} or \textit{Haptics}. Instead, our proposed interface enabled users to complete these tasks more efficiently. \textbf{LIMIT} resulted in significantly less distance traveled for \textit{Sounds} ($t(109)=2.24$, $p<.05$) and \textit{Haptics} ($t(54)=3.67$, $p<.001$). Along the same lines, \textbf{LIMIT} led to shorter interactions for \textit{Sounds} ($t(109)=3.1$, $p<.05$) and \textit{Haptics} ($t(54)=3.98$, $p<.001$). We conclude that interfaces which co-adapted to participants using \textbf{LIMIT} selected more helpful signals than \textbf{Bayes}. But we also recognize that the way in which \textbf{LIMIT} helped users differed from one interface to another. For the \textit{Lights} interface \textbf{LIMIT} improved the human's task reward without significantly changing the distance traveled or time taken. By contrast, for the \textit{Sounds} and \textit{Haptics} interfaces \textbf{LIMIT} and \textbf{Bayes} both obtained similar task reward, but \textbf{LIMIT} enabled users to reach this reward more quickly and efficiently.

\begin{figure}[t]
    \begin{center}
    \includegraphics[width=0.9\columnwidth]{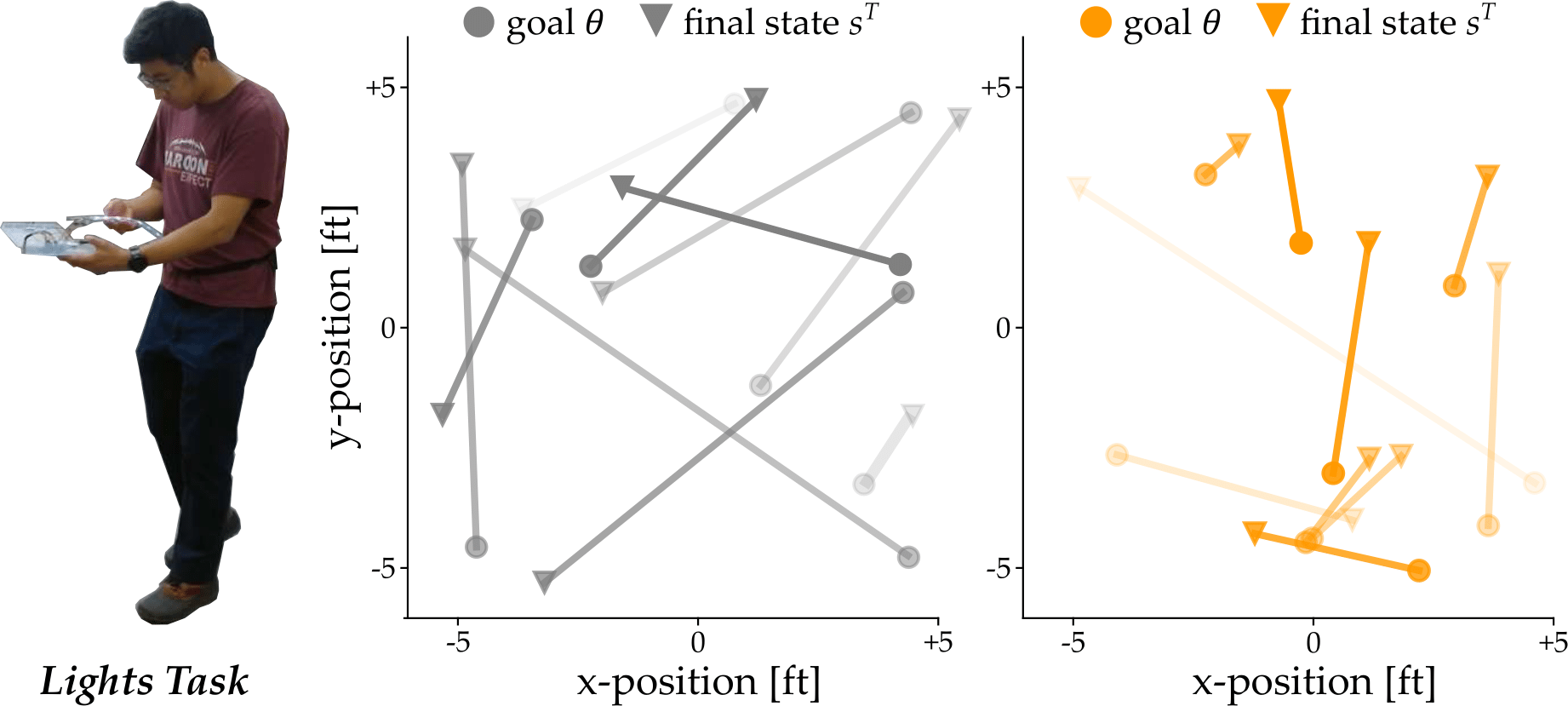}
		\caption{
  Example data from \textit{Lights}. (Middle) For one user we plot the error between their final position and the goal position as a function of interaction number for both \textbf{Bayes} and \textbf{LIMIT}. (Right) Ideally the final state should match the goal $\theta$. We visualize this same data by displaying the user's final $x$-$y$ position and the actual $x$-$y$ location of the goal. The user completed a total of $10$ interactions: lighter lines signify their first interactions and darker lines their final interactions. Error generally decreases over time with \textbf{LIMIT}.
  }
		\label{fig:lights}
	\end{center}
\end{figure}

To illustrate how \textbf{Bayes} and \textbf{LIMIT} affected the human's performance we highlight a \textit{Lights} example in \fig{lights}. Across $10$ repeated interactions the human walked to reach hidden goals $\theta$. When the interface used \textbf{Bayes} to select signals $x$, the human's error was roughly constant from one interaction to another. But under \textbf{LIMIT} this interface and human co-adapted so that the human could more accurately interpret the interface's signal after four interactions. 

We now turn to hypothesis \textbf{H2} and our Likert-scale survey in \fig{subjective}. Our survey items are listed in Appendix~\ref{sec:appendix}. To process the subjective results we first confirmed that each of the four scales were reliable (Cronbach's $\alpha>0.7$). We then grouped each multi-item scale into a combined score and performed paired $t$-tests to determine whether \textbf{LIMIT} received significantly higher scores than \textbf{Bayes}. Users reported that their performance \textit{improved} more with \textbf{LIMIT} ($t(32)=2.75$, $p<.05$), and they better \textit{understood} what the \textbf{LIMIT} interface was trying to convey ($t(32)=3.08$, $p<.05$). The participants also rated \textbf{LIMIT} as having more \textit{consistent} feedback than \textbf{Bayes} ($t(32)=2.58$, $p<.05$). When asked to explain their preferences, users commented that LIMIT ``\textit{was a lot more intuitive to adjust to},'' in part because it ``\textit{seemed more consistent}'' and maintained a distinct, one-to-one mapping from information $\theta$ to signals $x$.

\p{Discussion} The experimental results from our in-person user study suggests that \textbf{LIMIT} outperforms \textbf{Bayes} in an ensemble of interfaces, both preferentially and numerically. However, we note that the difference between performance (albeit significant) is not as large as one may expect. We emphasize that the \textbf{Bayes} algorithm \textit{has access to the human's intent} (i.e., their goal or reward function), giving it a significant advantage to \textbf{LIMIT}. However, as is evidenced empirically, users were able to adapt to signals produced by \textbf{LIMIT} more easily and achieved better results in higher-dimensional settings. {In more complex tasks, the performance gap between \textbf{LIMIT} and \textbf{Bayes} is more clear, even when \textbf{Bayes} has access to the human's intent (see Section \ref{adx:complex})}. 

\p{Summary} Our in-person user study evaluated \textbf{LIMIT} across three types of interfaces: audio feedback, visual feedback, and haptic feedback. The experimental results suggest that \textbf{LIMIT} learns to select meaningful and interpretable signals that help users complete their tasks. Across all objective and subjective metrics, we found that \textbf{LIMIT} scored as well as or better than a state-of-the-art baseline that has access to the task reward.

\section{Conclusion}

In this paper we introduced LIMIT, a co-adaptive approach to learn interface mappings from scratch. Learning interfaces is challenging because the way people respond to signals varies across tasks, users, and interface types. To address these challenges we hypothesized that interfaces should learn policies that maximize correlation between the human's actions and the interface's information. We derived a learning algorithm that updates the interface's signals in real-time to optimize for a tractable proxy of information gain. We then put LIMIT to the test across controlled simulations, an online survey, and in-person user studies. When compared to naive baselines and a state-of-the-art alternative with auditory, visual, and haptic interfaces, we found that LIMIT results in better task performance and higher subjective ratings.

\p{Limitations} LIMIT is a step towards robots that autonomously personalize their feedback for the current user. One advantage of LIMIT is that it does not need to know what task the human is trying to complete (i.e., what the human is using the signals for). {Our key assumption here is that --- no matter what task the human has in mind --- the human should respond in different ways to different hidden information $\theta$.} While maximizing information gain makes sense for the experiments presented in this paper, there are also settings where the human should maintain the \textit{same} actions even when the hidden information \textit{changes}. For example, imagine a driving scenario where the interface is communicating the location of nearby cars and pedestrians. The human is driving straight ahead, and should not change their actions if another car passes by or if a pedestrian is walking on the opposite sidewalk. When applied to this context, LIMIT may learn to cause the human driver to slow down or speed up, even though these changes are not necessary. {Our future work will explore how LIMIT can be combined with task-specific objectives to ensure that the signals are always necessary and meaningful. Our initial hypothesis here is that the task-specific reward function $R(\xi, \theta)$ could be incorporated within \eq{M12} so that LIMIT trains interfaces to simultaneously maximize interpretability and performance.}

\p{Future Works} 
A key application of LIMIT could be in sensory substitution, a scenario where neither the engineer nor the user has an informed prior over interface design. For example, brain-computer interfaces (BCI) have been connected with proprioceptive feedback \cite{deo2021effects}. Here the proprioceptive feedback (e.g., haptic stimulation) could be tuned using an approach like LIMIT to improve the accuracy of the BCI interface. Along similar lines, vibrotactile biofeedback interfaces \cite{chatterjee2007brain} could be made more effective using LIMIT to adjust the mapping between vibrotactile stimuli and user inputs. Moving forward, we see LIMIT as a step towards robots and interfaces that autonomously change their feedback to make the system more intuitive, understandable, and user-friendly.

%%%%%%%%%%%%%%%%%%%%%%%%%%%%%%%%%%%%%%%%%%%%%%%%%%%%%%%%%%%%%%%%%%%%%%%%%%%%%%%%%

\bibliographystyle{ACM-Reference-Format}
\bibliography{bibtex}

%%%%%%%%%%%%%%%%%%%%%%%%%%%%%%%%%%%%%%%%%%%%%%%%%%%%%%%%%%%%%%%%%%%%%%%%%%%%%%%%%

\appendix
\section*{Appendix}
\section{Likert Survey Tables} \label{sec:appendix}
\subsection{Online User Study (Section~\ref{sec:amt})} \label{sec:appendix-online-likert}
This section lists the questions on the Likert scale survey from our online user study in Section \ref{sec:amt}.
We organised questions into three scales (\textit{Intuitive}, \textit{Understand}, and \textit{Improve}) and tested their reliability using Cronbach's $\alpha > 0.7$. We then grouped each multi-item scale into a combined score and performed paired \textit{t}-tests to determine whether \textbf{LIMIT} received significantly higher scores than \textbf{Naive}. The results of the \textit{t}-tests are reported in Section~\ref{sec:amt}. Here we list the exact items and their corresponding scale.

\begin{table*}[h]

	\caption{Questions from our online user study in Section~\ref{sec:amt}}
	\label{table:likert-online}
	\centering
		\begin{tabular}{lcccc}
			\hline Questionnaire Item & Scale \bigstrut \\ \hline 
                \bigstrut[t]
            
            -- I thought the signals had a consistent pattern. & \multirow{2}{*}{\textit{Intuitive}}  \\  -- The signals seemed inconsistent or random. \bigstrut[b] \\ \hline  
                \bigstrut[t]
            
            -- By the end I could accurately predict the phone location. & \multirow{2}{*}{\textit{Understand}} \\ -- At the end I was still unsure what the interface was trying to convey. \bigstrut[b] \\ \hline  
                \bigstrut[t]

            -- I felt like my performance improved over time. & \multirow{2}{*}{\textit{Improve}} \\ -- It seemed like my performance stayed about the same. \bigstrut[b] \\ \hline  
		\end{tabular}

\vspace{-0.5em}

\end{table*}
\subsection{In-Person User Study (Section~\ref{sec:user-study})}
 \label{sec:appendix-inperson-likert}
This appendix lists the questions on the Likert scale survey from Section \ref{sec:user-study}. We organized the items into four scales (\textit{Intuitive}, \textit{Understand}, \textit{Consistent}, and \textit{Improve}) and tested their reliability using Cronbach's $\alpha > 0.7$. We then grouped each multi-item scale into a combined score and performed pair \textit{t}-tests to determine whether \textbf{LIMIT} received significantly higher scores than \textbf{Bayes}. The results of \textit{t}-tests are reported in Section~\ref{sec:user-study}. Here we list the exact items and their scale.

\begin{table*}[h]

    \caption{Questions from our in-person user study in Section~\ref{sec:user-study}}
	\label{table:likert-inperson}
	\centering
		\begin{tabular}{lcccc}
			\hline Questionnaire Item & Scale \bigstrut \\ \hline 
                \bigstrut[t]
            
            -- If I used the interface more, I think I would understand what it was trying to say. & \multirow{2}{*}{\textit{Intuitive}}  \\  -- Even if I kept practicing with this interface, I still don't think I would get it. \bigstrut[b] \\ \hline  
                \bigstrut[t]
            
            -- By the end I could understand what the interface was saying. & \multirow{2}{*}{\textit{Understand}} \\ -- At the end I was still unsure what the interface was trying to convey. \bigstrut[b] \\ \hline  
                \bigstrut[t]

            -- I thought the signals had a consistent pattern. & \multirow{2}{*}{\textit{Consistent}} \\ -- The signals seemed inconsistent or random. \bigstrut[b] \\ \hline  
                \bigstrut[t]
            
            -- I felt like my performance improved over time. & \multirow{2}{*}{\textit{Improve}} \\ -- It seemed like my performance stayed about the same. \bigstrut[b] \\ \hline  
		\end{tabular}

\vspace{-0.5em}

\end{table*}

\section{Network Architecture}\label{adx:arch}

In this section, we detail the specifics of the neural network architectures used throughout this paper. Nearly all networks in all experiments were multilayer-perceptrons (MLPs), but Table \ref{table:architecture} lists the architecture in detail. 

\begin{table*}[h]
    \caption{Neural Network Architecture in Simulations (Section \ref{sec:sims}) and User Studies (Section \ref{sec:user-study})}
	\label{table:architecture}
	\centering
		\begin{tabular}{lcccc}
			\hline Sim / User Study & Hidden Layers & Hidden Layer Size\footnotemark[1] & Activation Functions\footnotemark[2] & LR \bigstrut
			\\ 
                \hline  
                \bigstrut[t]
                1-DoF Sim & 2 & 8, 8 & \texttt{Tanh} & 0.01
                \\
                2-DoF Sim & 2 & 16, 64 & \texttt{ReLU}, \texttt{Tanh} & 0.001
                \\
                3-DoF Sim & 3 & 36, 96 & \texttt{ReLU}, \texttt{Tanh} & 0.001
                \bigstrut[b]
                \\
                \hline
                \bigstrut[t]
                1-DoF User Study & 2 & 8, 16 & \texttt{Tanh} & 0.01
                \\
                2-DoF User Study & 2 & 16, 32 & \texttt{Tanh} & 0.01
                \\
                3-DoF User Study & 3 & 36, 64 & \texttt{Tanh} & 0.01
                \bigstrut[b]
                \\
                \hline
                \bigstrut[t]
                Highway Sim & 3 & 36, 64 & \texttt{ReLU}, \texttt{Tanh}, \texttt{Sigmoid} & 0.001
                \\
                10-DoF Sim & 2 & 64, 128 & \texttt{ReLU}, 
                \texttt{Tanh}
                & 0.00025
                \bigstrut[b]
                \\
                \hline
   \end{tabular}

\vspace{-0.5em}

\end{table*}
\footnotetext[1]{The first number listed refers to the human and interface networks, the second number listed refers to the size of the decoder's hidden layers; these are much larger than the human and interface models.}
\footnotetext[2]{\texttt{Tanh} was always used on the interface neural network to restrict the signal output. Whenever another activation function is listed, it was used for the decoder and human model networks. For more detail, see our Github repository.}

Note that although the decoder networks used in this study were MLPs, other architectures could be used (such as gated recurrent units (GRUs)). Further, it is possible to use more exotic structures for LIMIT (such as Bayesian networks), but feed-forward networks were easier for us to use and demonstrate. For implementation specifics, see our Github repository.

\section{Additional Simulations}\label{adx:sim}
\subsection{Scenarios Requiring Near-Immediate Signal Adaptation}\label{adx:immed}

\begin{figure*}[t]
    \begin{center}
    \includegraphics[width=0.85\columnwidth]{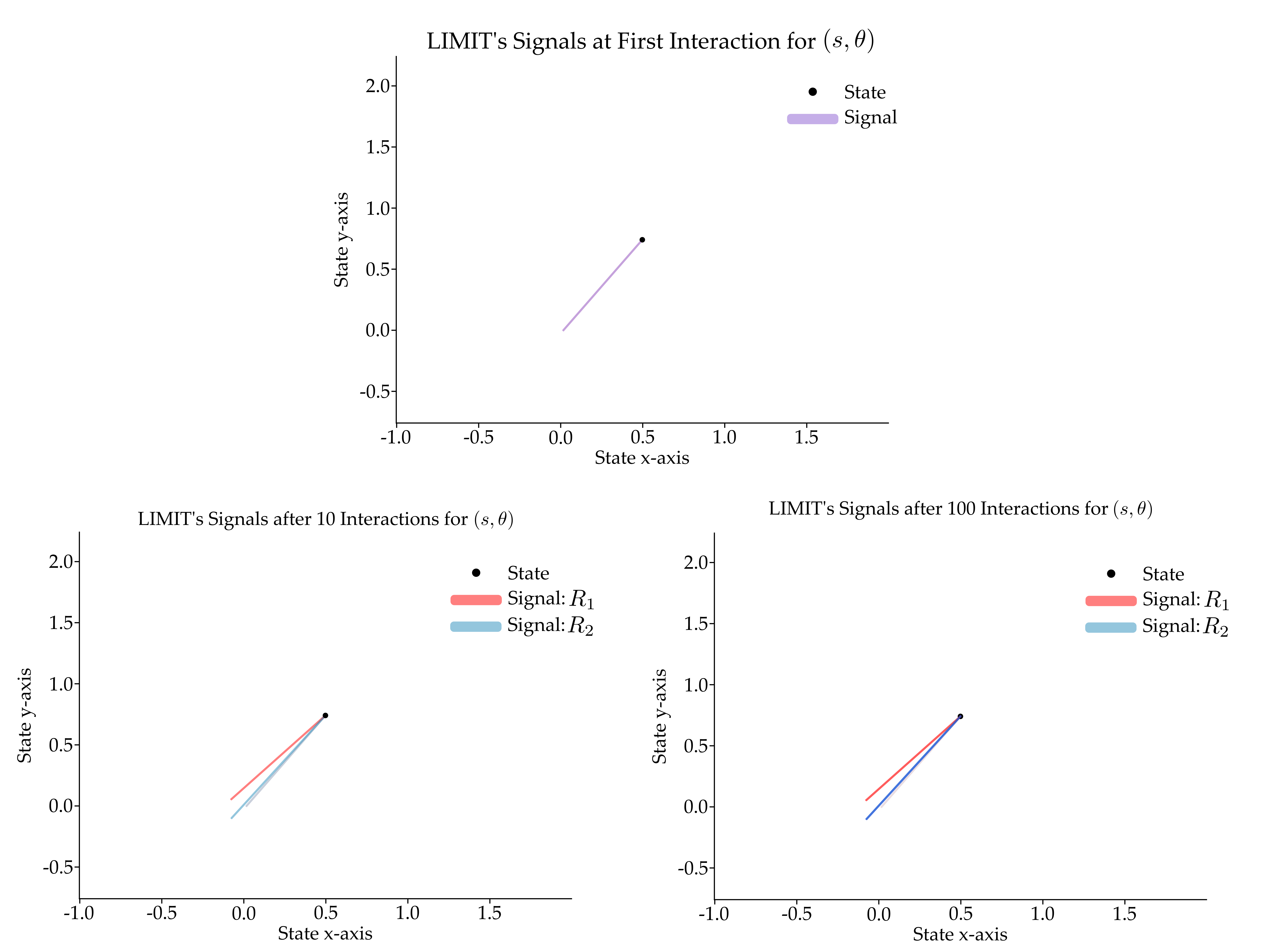}
		\caption{The change in LIMIT's signals over repeated interaction for different human policies in a 2-DoF environment. (Top) The initial signal for the first human model $\mathcal{H}_0$ for $\left(s, \theta\right)$, pretrained for one interaction. (Bottom Left) The signals \textit{plotted for each interaction} for the same instance of LIMIT, trained with new human models $\mathcal{H}_1$ and $\mathcal{H}_2$ for 10 interactions. Note that the signals are different than the first figure, and that the signals change slightly over repeated interaction as LIMIT adjusts to the new users' behavior. (Bottom Right) Signals produced by LIMIT for $\mathcal{H}_1, \mathcal{H}_2$ at $\left(s, \theta\right)$ after 100 interactions. Note that after pretraining, $\mathcal{R}_1$ and $\mathcal{R}_2$ are separate instances of LIMIT \textit{trained with different replay memory buffers and different optimizers}.}
		\label{fig:adx-change-in-signals}
	\end{center}
\end{figure*}

LIMIT is intended for settings where the human repeatedly interacts with an interface, and the interface can tune its signal over time. This may result in the system being less efficient within situations where the user requires near-immediate understanding of the interface's signals (e.g., understanding the interface at the first interaction). To better understand how quickly LIMIT personalizes its signals, we have conducted additional simulations using the same environment as Section \ref{sec:sims-2d}. We first pretrained an instance of LIMIT while interacting with a simulated human. We then sampled a random state-hidden information pair $(s, \theta)$ and plotted the signal $x$ LIMIT learned with the baseline human. Figure \ref{fig:adx-change-in-signals} (Top) shows the initial signal.

We next paired two copies of the pretrained LIMIT with two different simulated humans. Each of these simulated humans had a different policy than the baseline; the new human policies were randomly sampled. We then plotted the adapted signals that LIMIT learned after 10 interactions and 100 interactions with both of the new simulated humans. See Figure \ref{fig:adx-change-in-signals} (Bottom). Comparing the plots, we conclude that the interface’s learned signals have converged within 10 interactions: the signals the interface sends at 10 interactions are the same signals the interface uses after 100 interactions. Overall, Figure \ref{fig:adx-change-in-signals} suggests that LIMIT can personalize to new users in less than 10 interactions. 

\subsection{Variance in LIMIT's Signals Between Users}

\begin{figure*}[t]
    \begin{center}
    \includegraphics[width=0.7\columnwidth]{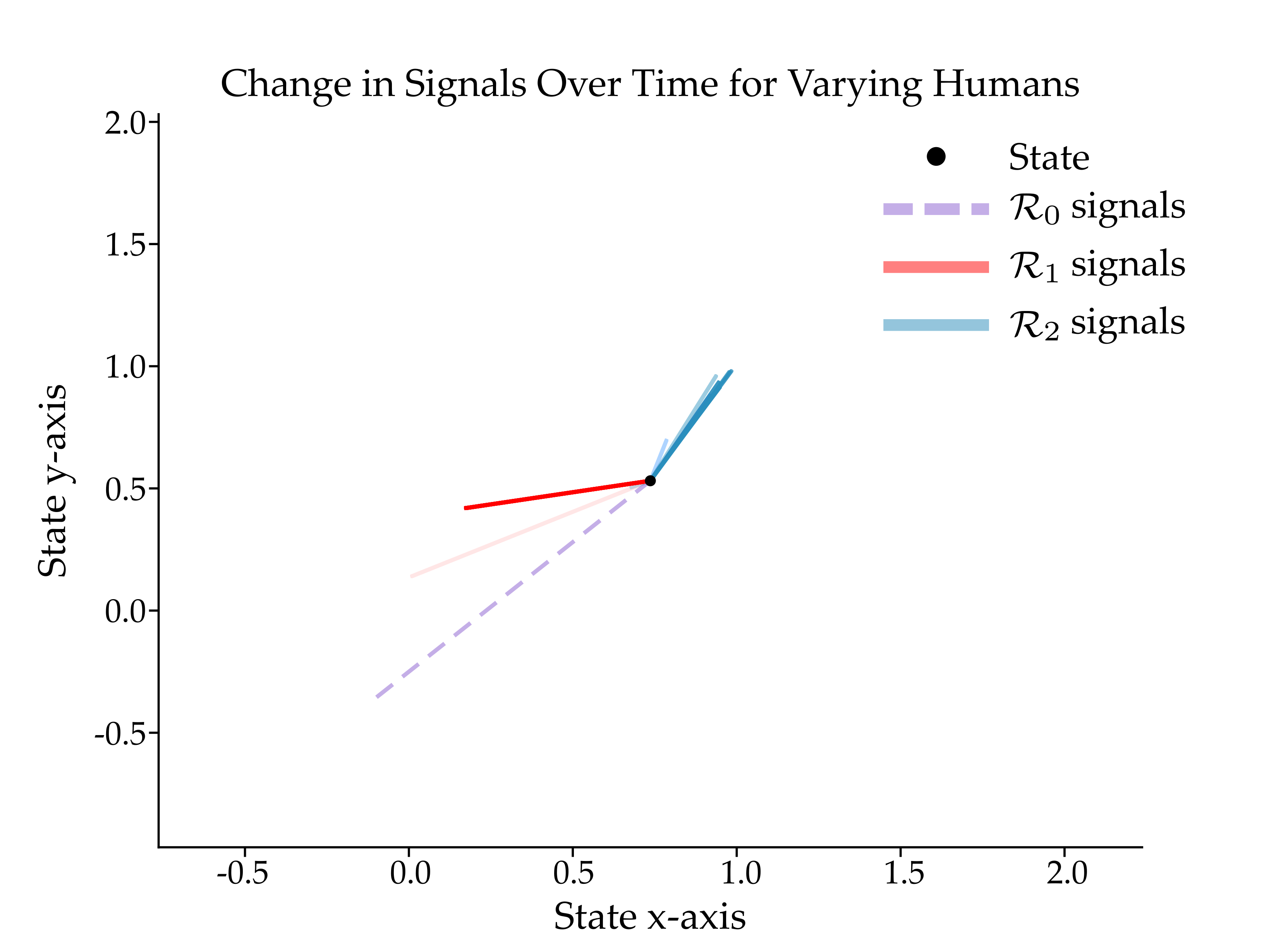}
		\caption{The change in signals over time for new humans $\mathcal{H}_1$ and $\mathcal{H}_2$ is clearly shown for an arbitrary yet particular $(s, \theta)$. Here, the signals are shown as vectors pointing away from the "state" point. Note that the signals' $x$- and $y$-axis correspond with the axis of the plot, i.e. a vector pointing along the $x$-axis with one unit of length would correspond to a signal of $\begin{bmatrix} 1 \  0 \end{bmatrix}^T$.}
		\label{fig:adx-change-in-signals-2}
	\end{center}
\end{figure*}

To investigate whether LIMIT produced different signals for different users, we conducted an additional simulation, similar to that of Appendix \ref{adx:immed}. Here, we pretrained an instance of LIMIT $\mathcal{R}_0$ on a human model $\mathcal{H}_0$ in our standard 2-DoF environment described in Section \ref{sec:sims-2d} (matching the "Lights" task from our in-person user studies). After pretraining LIMIT for several interactions with $\mathcal{H}_0$, we duplicated the instance of LIMIT and paired each with two new human agents with distinct policies $\mathcal{H}_1$ and $\mathcal{H}_2$. Then, after each interaction, we plotted the signals observed for each instance of LIMIT ($\mathcal{R}_1$ and $\mathcal{R}_2$) for a particular $(s, \theta)$. Figure \ref{fig:adx-change-in-signals-2} shows this clearly: LIMIT adjusts its signals for new users over repeated interaction, accommodating their distinct behaviors.

\end{document}